\pdfoutput=1

\documentclass[11pt]{article}
\usepackage{misc/eacl_2023}
\usepackage{times}
\usepackage{latexsym}
\usepackage[T1]{fontenc}
\usepackage[utf8]{inputenc}
\usepackage{microtype} %
\usepackage{todonotes}
\usepackage{amsmath}
\usepackage{amssymb}
\usepackage{cleveref}
\usepackage{booktabs}
\usepackage{lipsum}
\usepackage{enumitem}
\usepackage{xspace}
\usepackage{listings}
\usepackage{soul} %
\usepackage{caption}
\usepackage{transparent} %
\usepackage{tcolorbox} %
\usepackage{mdframed} %

\crefname{lstlisting}{listing}{listings}
\Crefname{lstlisting}{Listing}{Listings}
\crefname{equ}{equation}{equations}
\Crefname{equ}{Equation}{Equations}

\DeclareCaptionType{equ}[Equation][List of equations]

\usepackage{courier}
\usepackage{marginnote}

\definecolor{TodoColor}{rgb}{1,0.7,0.6}

\sethlcolor{TodoColor}

\definecolor{DocumentLinkColor}{rgb}{0.4,0.6,0.3}
\newcommand{\hrefEmail}[2]{\href{mailto:#1}{\color{black}{#2}}}

\newcommand{\whitezero}{\textcolor{white}{0}}
\newcommand{\nominus}{\text{}\hspace{1mm}}
\newcommand{\nospacestar}{$\star$\hspace{-3.2mm}\text{}}

\newcommand*\ttjustify{%
  \fontdimen2\font=0.4em%
  \fontdimen3\font=0.2em%
  \fontdimen4\font=0.1em%
  \fontdimen7\font=0.1em%
  \hyphenchar\font=`\-%
}

\definecolor{ethblue}{rgb}{0,0.1,0.4}
\newcommand{\ethletter}{\hspace{-0.5mm}\text{
    \fontfamily{phv}\fontseries{bx}\fontsize{7}{\baselineskip}\selectfont
    \textit{\textbf{\color{ethblue}{E}}}}
}
\newcommand{\msftletter}{\hspace{-0.5mm}\text{
    \fontfamily{phv}\fontseries{bx}\fontsize{7}{\baselineskip}\selectfont
    \textbf{M}}
}
\newcommand{\tudletter}{\hspace{-0.5mm}\text{
    \fontfamily{phv}\fontseries{bx}\fontsize{7}{\baselineskip}\selectfont
    \textbf{D}}
}

\newmdenv[
  linecolor=black,
  linewidth=1.2pt,
  topline=false,
  bottomline=false,
  rightline=false,
  innertopmargin=0mm,
  innerbottommargin=-0.5mm,
  skipabove=1.1\topsep,
  skipbelow=0.5\topsep,
]{quotebox}

\newcommand{\researchquestion}[2]{
    {
    \vspace{-1.2mm}
    \begin{tcolorbox}[colback=white!0,boxrule=0.9pt,right=2mm,left=2mm,bottom=2mm,top=2mm]
        \fontsize{10.5pt}{10.5pt}\selectfont
        \vspace{-1mm} \textbf{#1}: #2
        
        \vspace{-0.8mm}
    \end{tcolorbox}
    }
    \vspace{-1mm}
}

\newcommand{\textexample}[1]{
    \vspace{2mm}
    \hspace{-4mm}
    \begin{minipage}{1.04\linewidth}
    \begin{quotebox}
        \setlength{\parindent}{0cm}
        #1
    \end{quotebox}
    \end{minipage}
}

\newcommand{\smallmath}[1]{
#1
}

\usepackage{todonotes}

\title{Poor Man's Quality Estimation:\\ Predicting Reference-Based MT Metrics Without the Reference}

\newcommand{\quadtop}{\hspace{7mm}}

\author{
    Vilém Zouhar$^{\ethletter}$ \quadtop
    Shehzaad Dhuliawala$^{\ethletter}$ \quadtop
    Wangchunshu Zhou$^{\ethletter}$ \quadtop
    Nico Daheim$^{\tudletter}$ \\
    \textbf{
    Tom Kocmi$^{\msftletter}$ \qquad
    Yuchen Eleanor Jiang$^{\ethletter}$ \qquad
    Mrinmaya Sachan$^{\ethletter}$
    } \\ \text{} \\
  $^{\ethletter}$\textbf{ETH Zürich} \quad $^{\msftletter}$\textbf{Microsoft} \quad
  $^{\tudletter}$\textbf{TU Darmstadt} \\
  \texttt{\{\hrefEmail{vzouhar@ethz.ch}{vzouhar},\hrefEmail{msachan@ethz.ch}{msachan}\}@ethz.ch}
}

\begin{document}
\maketitle

\begin{abstract}
Machine translation quality estimation (QE) predicts human judgements of a translation hypothesis without seeing the reference.
State-of-the-art QE systems based on pretrained language models have been achieving remarkable correlations with human judgements yet they are computationally heavy and require human annotations, which are slow and expensive to create.
To address these limitations, we define the problem of \emph{metric estimation} (ME) where one predicts the automated metric scores also without the reference.
We show that even without access to the reference, our model can estimate automated metrics ($\rho {=} 60\%$ for sentBLEU, $\bar{\rho} {=} 51\%$ for other metrics) at the sentence-level.
Because automated metrics correlate with human judgements, we can leverage the ME task for pre-training a QE model.
For the QE task, we find that pre-training on TER is better ($\rho{=}23\%$) than training for scratch ($\rho{=}20\%$).
\end{abstract}

\hspace{-3mm}
\begin{minipage}[c]{4.5mm}
\includegraphics[width=\linewidth]{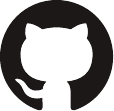}
\end{minipage}
\hspace{0mm}
\begin{minipage}[c]{0.7\textwidth}
\fontsize{8pt}{8pt}\selectfont
\href{https://github.com/zouharvi/mt-metric-estimation}{\texttt{github.com/zouharvi/mt-metric-estimation}}
\end{minipage}

\section{Introduction}

Quality estimation (QE) is often used in machine translation (MT) production pipelines where we need to make decisions based on the quality of an MT output but where the reference is unavailable \citep{specia2020findings,specia2021findings}.
For example, QE is used in translation companies to decide whether to send a specific MT output for post-editing to a human translator or whether to use it directly \citep{tamchyna2021deploying,zouhar2021backtranslation,murgolo2022quality}.
In this scenario, an accurate QE system has the potential to save expensive translator effort.
However, training QE models usually requires human-annotated judgements of the translation quality \citep{specia2013quest,rubino2020intermediate,rei2020comet}.
These human annotations are scarce and costly, especially for low-resource language directions, and may need to be replicated for new MT systems and domains \citep{de2014machine}

We investigate if automated MT metrics can be used to reduce the cost of learning QE models.
Automated metrics can be run with no additional costs and can be used to generate large amounts of QE data.
If these metrics correlate well with human judgements, this larger data can be used as a partial substitute for human data during training.
Data augmentation and synthetic QE data via automated metrics has already been explored \citep{heo2021quality,baek2020patquest,cui2021directqe}, though never in the pre-training \& fine-tuning fashion.

\begin{figure}
\centering
\includegraphics[width=0.9\linewidth]{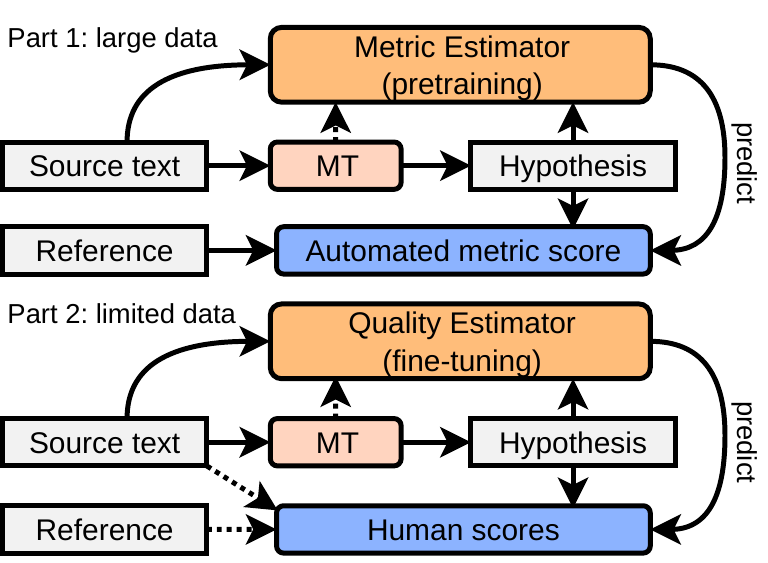}
\caption{Pipeline for pre-training on automated metrics (top) and fine-tuning on limited quality estimation data (bottom). Dotted lines are optional dependency.}
\label{fig:pipeline}
\vspace{-4mm}
\end{figure}

Our work is guided by a simple intuition.
Intuitively, human judgements can be thought of as functions depending on the target sequence and (optionally) the source and reference(s): $f_\textsc{human}([s], h, [r])$.
The task of QE is to model $f_\textsc{human}$ based only on the source $s$ and hypothesis $h$.
Because the function arguments of $f_\textsc{human}([s], h, [r])$ resemble those of automated metrics for MT: $f_\textsc{metric}([s], h, r)$, we can use the automated metrics to guide learning  of the human quality judgements which are hard to obtain and replicate.
Generating automated metric scores is limited only by the amount of parallel data which is more abundant.
Because $f_\textsc{metric}$ correlates with human judgement $f_\textsc{human}$ \citep{ma2019results,mathur2020results}, we can start by estimating $f_\textsc{metric}$ and only later fine-tune to $f_\textsc{human}$.
We refer to estimating $f_\textsc{metric}(s, h, r)$ as \textbf{\ul{m}etric \ul{e}stimation (ME)} as a parallel task to \textbf{\ul{q}uality \ul{e}stimation (QE)}.

We illustrate our idea of pre-training on the automatic metrics and fine-tuning on human assesments in \Cref{fig:pipeline}.
Our model uses a BiLSTM with the source and hypothesis as the input (with several more features like the decoder confidence and hypothesis variance) to output a single number (metric or quality score).

Experimentally, we find the idea of mitigating data limitation for QE with ME pretraining challenging.
Thus, we structure our investigation around a set of research questions. First, we try to establish that it is possible to robustly predict automated metrics and explore the associated data requirements.
Then, concerned with the application and deployment of the ME model, we also check how transferrable the model is between different MT systems.
We break the research down into the following questions.

\researchquestion{RQ1}{Can automated reference-based MT metrics be reliably predicted without the reference? \\
\textbf{RQ2}: What is the effect of data size on predicting automated reference-based MT metrics? \\
\textbf{RQ3}: To what extent does transferring between MT systems damage metric estimation models? \\
\textbf{RQ4}: Can metric estimation be used as a pre-training step for quality estimation?
}

We answer the research questions with experiments on the English $\rightarrow$ German language direction and replicate the main findings on 12 language pairs in total in \Cref{sec:other_langs}.
We confirm that sentBLEU is predictable with $\rho = 60.4\%$ sentence-level Pearson's correlation and other metrics with $\bar{\rho} = 51.3\%$ (\textbf{RQ1}, \Cref{sec:results_baselines}).
Authentic parallel data is needed for ME models but this can be alleviated by using more hypothesis from beam search  (\textbf{RQ2}, \Cref{sec:data_size}).
It is possible to train the ME system on one MT system and then use it on a different MT system with only a slight loss in performance (\textbf{RQ3}, \Cref{sec:model_transfer}).
We find that pre-training on Translation Edit Rate {(TER)} \citep{snover2006study} leads to better results than training on the QE data directly, though this approach does not outperform the state-of-the-art in QE (\textbf{RQ4}, \Cref{sec:finetuning_human}).

\begin{figure}[htbp]
\centering

\includegraphics[width=\linewidth]{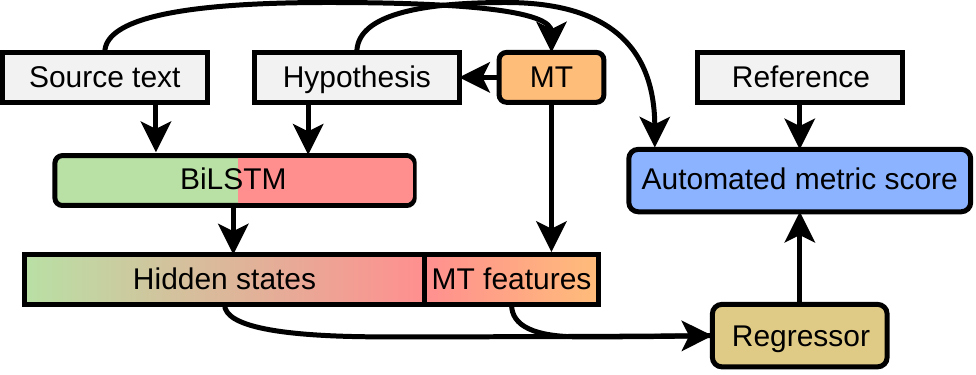}

\caption{Metric/quality estimation model architecture.}
\label{fig:model}
\end{figure}

\section{Our Metric Estimation Model}
\label{sec:system}

\paragraph{Notation.}

Given a source sentence $s$ and its translation, $h$ which is an output of a MT system ($h = \text{MT}(s)$), we build a regressive ME model which outputs a numeric score that %
is close to the
output of an automated metric that is further dependent on the reference: $f_\textsc{metric}(s, h, r)$.
We distinguish two cases based on our level of access to the MT system: {\bf blackbox} setting (where we assume access only to the MT system output) and {\bf glassbox} setting (where we have access to entire MT model).
In the later case, we may leverage features from the MT model to improve the ME capabilities.

\paragraph{Model.}
Our main model for ME/QE, shown in \Cref{fig:model}, starts with Byte-Pair-Encoding \citep{gage1994new,Sennrich2016NeuralMT} the source $s$ and hypothesis $h$.
It is followed by BiLSTM on top of concatenated source and hypothesis with a separator ($s\texttt{[SEP]}t$).
The last hidden state (denoted as $\text{BiLSTM}_{-1}$) is extracted and fused together via concatenation with the internal MT system and other features (see the following list).
This is then used in a simple feed-forward layer ($\text{FFNN}$) to generate a single score number:
\smallmath{\begin{align*}
\text{source}\,\, s, \text{hypothesis} \,\, h := \text{MT}(s) \hspace{-4cm} \\
\Phi &= \text{BiLSTM}_{-1}( \text{BPE}(s\texttt{[SEP]}h)) \\
\Psi &= \text{Features}(\text{MT}_\text{conf.}(s, h), s, h) \\
\text{ME}_\text{all}(s, h) &= \text{FFNN} [\Phi, \Psi] \\
\text{ME}_\text{text}(s, h) &= \text{FFNN} [\Phi]
\end{align*}}

The first ME model is glassbox and the second is blackbox.
In the first case, we utilize hand-crafted features and also those from the MT system (function $\text{features}$).
Both of these models are optimized with mean-squared error against a particular metric.
That is, we train separate models for each target metric (COMET, ChrF, BLEURT, sentBLEU, METEOR, TER) or human judgements.
\smallmath{
\begin{gather*}
\mathcal{L} = \frac{1}{n} \sum (\text{ME}(s_i, h_i) - f_\textsc{METRIC}(s_i, h_i, r_i))^2
\end{gather*}
}

\noindent The additional features $\Psi$ are:
\begin{itemize}[noitemsep,topsep=1mm]
\item Decoder confidence (prob and logprob).
\item Source and target lengths and their relation. This is included as the distribution of errors may be different for various sentence lengths.
\item Average distance and variance between hypotheses as measured by an automatic metric.
\end{itemize}

\paragraph{Decoder confidence.}
Low probability MT outputs have overall lower quality \citep{specia2018quality,yankovskaya2018quality,fomicheva2020unsupervised}.
The decoder confidence is the hypothesis probability as defined by the model $\prod_i p(t_i| t_{<i}, s)$ which is in practice usually computed in the logspace $\sum_i \log p(t_i| t_{<i}, s)$.\footnote{The justification for using both $\text{conf}_t$ and $\exp(\text{conf}_t)$ is that the non-linear transformation improves correlation.}

\paragraph{Hypotheses variance.}

Intuitively, there are many ways to generate a wrong translation but only a few correct ones \citep{xu2011minimum}.
Similar to \citet{fomicheva2020unsupervised}, we hypothesise that larger variance between the hypotheses correlates negatively with quality.
We therefore use the distances between hypotheses as features for our system.
Specifically, as shown in \Cref{fig:features_hyp_space} and formalized with the following, we use the mean distance and also the variance between distances as features.
We first consider distances from the current hypothesis to be estimated (\texttt{H1}) to all other hypotheses, and then all hypothesis pairs.
\smallmath{\begin{gather*}
\text{Avg or Var}(\{ \textsc{sentBLEU}(h_1, h_j) | h_j \in H \}) \\
\text{Avg or Var}(\{ \textsc{sentBLEU}(h_i, h_j) | h_i, h_j \in H, i \neq j \})
\end{gather*}}

\begin{figure*}[htbp]
\centering
\includegraphics[width=\linewidth]{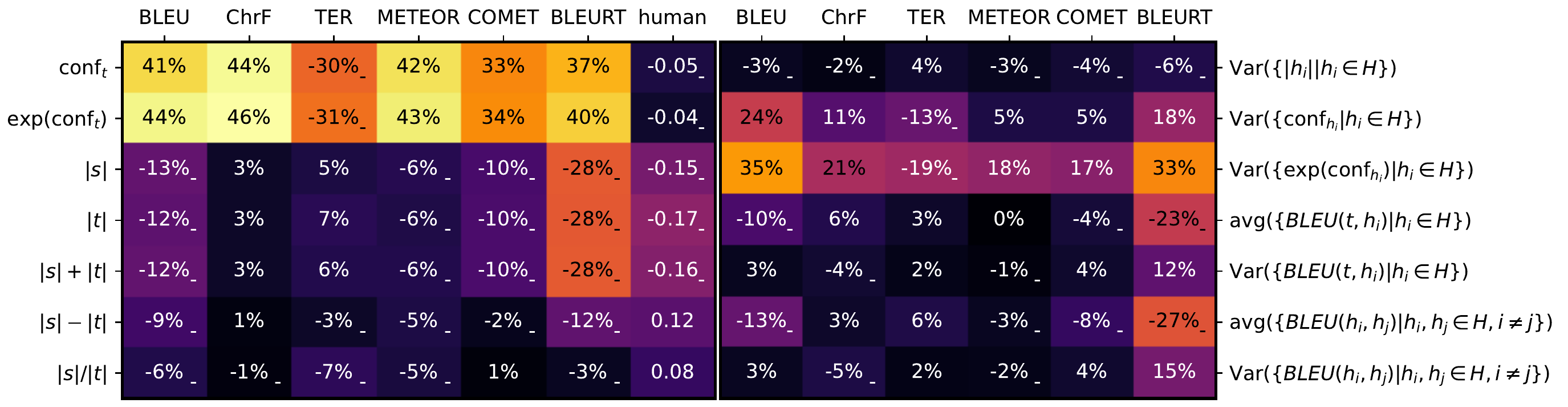}
\caption{Feature correlations on 500k English $\rightarrow$ German sentences with reference-based metrics and human judgement. Colour is based on absolute values to show contained relevant information. Cells with negative correlations are marked with ``-''.}
\label{fig:feature_corr}
\vspace{-3mm}
\end{figure*}

\begin{figure}[htbp]
\vspace{2mm}
\includegraphics[width=\linewidth]{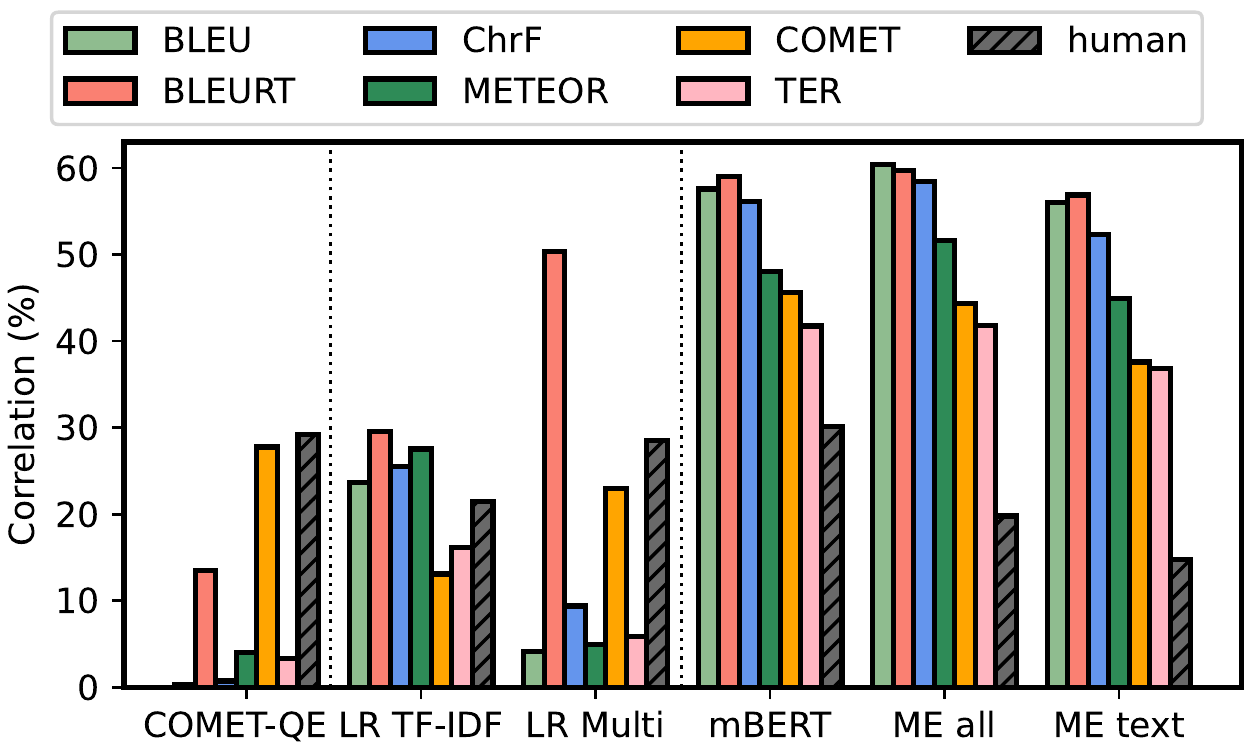}
\caption{Correlations with metrics and human judgement of baseline and main metric estimation (ME) models. Each bar is a separate model trained to predict a particular metric or human judgement. Individual features are shown in \Cref{fig:feature_corr}.}
\label{fig:baseline_comparison}
\end{figure}

\paragraph{Baselines.}

We use multiple baselines for comparison in the ME task.
Apart from the individual features, they are all optimized to minimize the MSE loss with a specific target metric.
\begin{itemize}[noitemsep,topsep=1mm]
\item Linear regression on TF-IDF features (with limited max features, see \Cref{sec:model_details}): $\,\text{Lin.Reg.}\big[\text{TF-IDF}(s, h) \big]$
\item Linear regression on all text \& MT features: \\
$\text{}\quad\text{Lin.Reg.}[\Psi]$
\item Fine-tuned mBERT \citep{devlin2019bert} with identical regression head architecture on top of last layer \texttt{[CLS]} hidden state\footnote{See \Cref{sec:model_details,sec:reproducibility} for details.}:
$\text{}\,\text{FFNN} \big[\text{mBERT}(s\texttt{[SEP]}h) \big]$ 
\end{itemize}

We also use an off-the-shelf QE model used in WMT21 QE task \texttt{wmt21-comet-qe-mqm} \citep{rei-EtAl:2020:WMT}.\footnote{This model was better than \texttt{wmt21-comet-qe-da}. Note the difference between the automated metric {COMET} and the QE system {COMET-QE}.}
We do not fine-tune the model to the available data, but since we use correlation as an evaluation metric, the mean is subtracted and the output rescaled to unit variance, same as human judgements.

\paragraph{Automated metrics.}
For ME we use the following MT metrics.
sentence-level implementation of BLEU \citep{papineni2002bleu} and ChrF \citep{popovic2015chrf} from SacreBLEU \citep{post2018call}, TER \citep{snover2006study} and METEOR \citep{banarjee2005} work with lexical or character-level units, commonly in word or character n-grams.
COMET \citep{rei2020comet} uses pre-trained encoders to evaluate the hypothesis at a deeper level.
BLEURT \citep{sellam2020bleurt} is another learned metric for text generation which uses pseudo-label.
While automated metrics usually yield only sentence-level scores, QE is done for multiple levels: word, phrase and sentence.
However, because of the automatic metric restrictions, we also focus on sentence-level QE in this work.

\section{Experiment Setup}

\paragraph{Pipeline \& data.}
We start by translating 500k English$\rightarrow$German sentences of the WMT14 dataset \citep{wmt14} and computing the automated metrics of these translations.\footnote{We are not limited by the relatively small size of this dataset because we are considering only its subset and study the effect of available data size in \Cref{sec:data_size}.}
We use a pre-trained WMT19 model by \citet{ng2019facebook}.

For human scores, we use the train data of WMT21 Sentence-Level Quality Estimation Shared Task \citep{specia2021findings} which contains 14k human-direct-assessment annotated segments (5k unique).
For the human score prediction, we do not have access to the features in the hypothesis space (because the hypotheses were not generated by the MT system to which we have access) and use forced decoding of the pre-trained model to get a confidence estimate.
Note that since the data comes from a different MT system, there is a distribution mismatch which may negatively influence the results.
We address this in \Cref{sec:model_transfer}.

We refer to the two datasets as ME and QE, respectively and show the distribution of automated metric scores and human judgements in \Cref{fig:metric_violin}.
We perform ME on both but QE only on the latter because of the human annotations availability.

\paragraph{Evaluation.}

We evaluate ME model performance with Pearson's coefficient with the target metric on a dev set of 10k sentences from WMT14 on the segment-level.
Note that we care about the \emph{magnitude (absolute value) of the correlation} and not whether it is positive or negative.
For example, TER is expected to correlate negatively with human ranking because higher TER means more errors while higher human scores mean higher translation quality.
Correlations with human judgement are evaluated on 1k WMT21 Sentence-Level QE data.
Specifically, we estimate human z-scores which were computed per-annotator.\footnote{Z-score of a variable has zero mean and unit variance. They are possibly unbounded but on the other hand, slightly alleviate the effect of individual annotator differences.}

\section{Results}
\label{sec:results}

This section first studies single-feature baselines (\Cref{sec:individual_features}) and then the possibility of robust ME model (\Cref{sec:results_baselines}) and data size requirements (\Cref{sec:data_size}).
The model is then checked on a different MT system output to see transferability between systems and architectures (\Cref{sec:model_transfer}).
Fine-tuning and evaluation on human data (QE) is done in \Cref{sec:finetuning_human}.
A natural follow-up experiment on using joint prediction to improve the ME model is documented in \Cref{sec:joint_prediction}.

\subsection{Feature analysis}
\label{sec:individual_features}
We show the correlations between individual features and metrics in \Cref{fig:feature_corr}.
An immediate observation is that confidence-based features correlate much more with automatic metrics than the other features.
Some metrics (sentBLEU and COMET) and especially human z-scores are highly correlated with the source and target lengths.
As a negative result, very few of the hypothesis space metrics correlate highly, with an exception of $\rho(\text{sentBLEU}, \text{Var}(\text{sentBLEU}(H)))$.
We still use all features later on because (1) despite low individual correlations, they may still be useful in combination or for the full model, whose input is the text, and (2) we did not encounter any overfitting issues.

\begin{table}[htbp]
\centering
\resizebox{\linewidth}{!}
{
\begin{tabular}{cccccc}
\toprule
\hspace{-1mm}\textbf{sentBLEU}\hspace{-1mm} &
\hspace{-1mm}\textbf{BLEURT}\hspace{-1mm} &
\textbf{ChrF} &
\hspace{-2mm}\textbf{METEOR}\hspace{-1mm} &
\hspace{-1mm}\textbf{COMET}\hspace{-2mm} &
\textbf{TER} \\
\midrule
11.1\% & 16.5\% & 12.3\% & -12.0\% & 11.5\% & 34.6\% \\
\bottomrule
\end{tabular}
}
\caption{Pearson correlations between human judgement (human z-scores) and automated metrics.}
\label{tab:metric_corr}
\end{table}

We show the correlations between the automated metrics and humans in \Cref{tab:metric_corr}.
For most of them, the correlations are very low.
Outliers with the highest correlation are BLEURT and COMET, which are known to be strong-performant metrics \citep{kocmi2021ship}.
One of the reasons is that they were specifically trained to correlate well with humans.

\subsection{Metric estimation performance}
\label{sec:results_baselines}

We show the baselines with comparison to the main models trained only on the target data (either WMT News or WMT QE) in \Cref{fig:baseline_comparison}.
Every bar is a separate model trained to predict a specific metric and the bar magnitude shows its correlation.
The systems are described formally in \Cref{sec:system}.
Notably, \emph{ME text} has access to only the source and hypothesis texts while \emph{ME all} in addition fuses in extra hand-crafted features.

A simple linear regression based on features from \Cref{fig:feature_corr} is able to achieve $>40\%$ correlations with automated metrics and ${\sim}13\%$ correlation with human judgement.
These features seem important as demonstrated by the comparatively lower correlations of a TF-IDF featurizer.
This is also documented by the difference between \emph{ME all} and \emph{ME text}.
The former model consistently outperforms \emph{ME text}, possibly because it has access to all the extra features while the latter model only works with the source and target texts.
The pre-training of the \emph{mBERT} model on language modelling helps only marginally, given that it performs only slightly better than \emph{ME text}.
The {COMET-QE} model almost does not correlate with automated metrics at all apart from its related reference-based metric {COMET}.
Of all models, it also correlates the most with human judgement. 

\researchquestion{RQ1}{Best-performing metric estimation models (mBERT, ME all, ME text) achieve sound correlations with automated metrics ($60.4\%$ with sentBLEU, $\sim 51.3\%$ for others).}

We can interpret this correlation with respect to the individual features performances ($\max \rho = 46\%$) which shows that the model was able to learn more predictive patterns.
On the selected datasets, the ME task is easier than the QE task where our models have a consistently lower performance.

\begin{figure}[htbp]
\includegraphics[width=\linewidth]{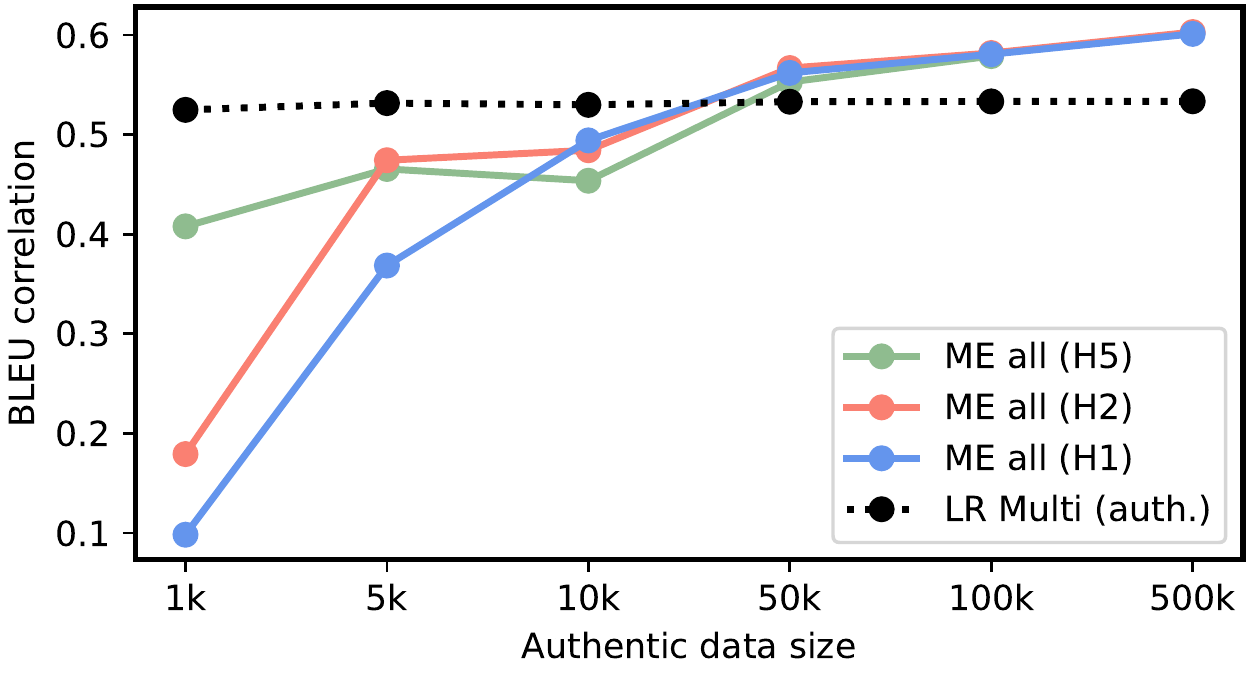}
\caption{Model dependency on training data size. H2 and H5 expand the data by top-2 and top-5 decoder hypotheses, respectively. Note the non-linear x-axis.}
\label{fig:limited_data}
\end{figure}

\subsection{Data requirements for ME}
\label{sec:data_size}

We are interested in how much data we need to train the ME models.
This is a practical question that helps us understand the behaviour and requirements of the models.

Naturally, models utilizing just a handful of continuous non-trainable features require much less data than models whose inputs are the raw texts.
This is demonstrated in \Cref{fig:limited_data} where the linear regression gains very little even if 500$\times$ more data is used.
For the main ME model, a larger amount of data is required (line H1).
So far we have been considering only the highest-scored hypothesis (in terms of decoder score) among 5 generated by beam search: $(s, h_1)$.
In a low-resource scenario, it may be beneficial to create more pairs from the hypotheses provided by the beam-search decoder: $\{(s, h_i)|h_i \in H\}$.
Through this hypothesis expansion, we may obtain more parallel data in data-restricted settings.
Again we show the results in \Cref{fig:limited_data} with a substantial gain of the model trained on expanded data (H2) over just the top hypothesis (H1).
This effect is quickly diminishing with larger data ($\geq$ 10k) but for lower data remains a useful tool.

\researchquestion{RQ2}{The metric estimation model requires $\sim$500k parallel sentences and the associated metric scores before the performance starts to plateau. For very low-resource scenarios, it is possible to reduce the ME model data requirements by utilizing multiple hypotheses for a single source sentence.}
    
\begin{table}[htbp]
\centering
\resizebox{\linewidth}{!}{
\begin{tabular}{lcccc}
\toprule
& \multicolumn{2}{c}{\textbf{ME All}} & \multicolumn{2}{c}{\textbf{ME Text}} \\
\textbf{Model} & \textbf{Transfer} & \textbf{Auth.}& \textbf{Transfer} & \textbf{Auth.} \\
\midrule
Original & \multicolumn{2}{c}{60.4\%} & \multicolumn{2}{c}{56.1\%} \\
W16Conv. \hspace{-1cm} & 54.2\% & 57.3\% & 48.4\% & 51.0\% \\
W16Trans. \hspace{-1cm} & 54.2\% & 57.1\% & 47.9\% & 50.7\% \\
W17Conv. \hspace{-1cm} & 58.7\% & 60.8\% & 55.1\% & 55.7\% \\
T5 & 19.8\% & 72.9\% & 43.8\% & 70.9\% \\
\bottomrule
\end{tabular}
}
\caption{sentBLEU estimation correlation for texts translated by different models than WMT19 Transformer (original).}
\label{fig:model_transfer}
\end{table}

\subsection{Generalization of ME across MT systems}
\label{sec:model_transfer}

In this section, we examine whether our ME model overfits on the specific errors the used MT system is doing or whether it generalizes and is applicable to also other MT systems.
This is important in deployment so that the possible ME/QE model is not dependent on a specific MT system.

We translate the same data using the following English $\rightarrow$ German models and store the decoder features:
T5-small \citep{raffel2020exploring}, WMT16 Convolutional \citep{wu2019pay}, WMT16 Transformer \citep{ott2018scaling}, WMT17 Convolutional \citep{gehring2017convolutional}.
We then run the ME model trained on the outputs of the original WMT19 system to predict metric scores for the outputs of these models.
We show the results in \Cref{fig:model_transfer}, for both text-only and feature-enriched models.
The evaluation of translations by different MT systems shows a varying decrease in correlation with the automated metrics.
For most systems, the drop was $\sim$2--3\%, which means that the metric estimator generalizes well across MT systems.
However, an exception is using a completely different model, the prompted T5 LM, for which the transfer mostly failed.
When training metric estimators on the T5 translated data, they achieved $\sim$71\% correlation but when models trained on outputs of a different MT system were used, the correlation dropped to 19.8\% and 43.8\% for \emph{ME All} and \emph{ME Text}, respectively. 
One of the reasons may be vastly different extra features, as documented by the noticeably higher correlation for the \emph{ME Text} model.

\researchquestion{RQ3}{It is possible to train a metric estimation model on sentences produced by MT$_1$ and use it to estimate metric scores of MT$_2$ with a drop in correlation of $\sim 3\%$.}
\vspace{-4mm}

\subsection{From ME to QE}
\label{sec:finetuning_human}

This section verifies empirically whether pre-training on ME helps on the QE task.
For this, we use models trained on estimating automatic metrics and either:
correlate their outputs with human judgements (zero-shot), or
fine-tune them to predict the human judgements directly (fine-tuning).
The fine-tuning was done with the same setup as in \Cref{sec:model_details} with all model parameters updated.

The results in \Cref{tab:transfer} show that fine-tuning definitely improved the performance over zero-shot.
However, only {TER} was able to outperform training on z-scores from scratch.
This can be attributed to it being the only metric with reasonable absolute correlation in the zero-shot.
Notably, {COMET} was better than the rest of the automated metrics with the worst being {sentBLEU}.
Despite the fine-tuning, we were not able to construct a QE system that would outperform the standard baseline of {COMET-QE} ($\rho$ = 29.2\%).

\begin{table}[htbp]
\centering
\begin{tabular}{lcc}
\toprule
\textbf{Pre-train metric} & \textbf{Zero-shot} & \textbf{Fine-tuning} \\
\midrule
\textbf{sentBLEU} & -3.2\% & -0.9\% \\
\textbf{BLEURT} & \whitezero{}5.9\% & -5.2\% \\
\textbf{ChrF} & -6.9\% & \whitezero{}6.5\% \\
\textbf{METEOR} & -4.2\% & \whitezero{}6.0\% \\
\textbf{COMET} & \nominus\whitezero{}1.1\% & 10.4\% \\
\textbf{TER} & -12.5\% & 22.8\% \nospacestar \\
\cmidrule{2-3} 
\textbf{human} & \multicolumn{2}{c}{19.8\%}\\
\bottomrule
\end{tabular}
\caption{Correlations with human judgement (z-scores) from models (with extra features fusion) which were pre-trained on metric estimation. Only the magnitude (absolute value) of the correlation is important.}
\label{tab:transfer}
\vspace{-4mm}
\end{table}

\vspace{-2mm}
\researchquestion{RQ4}{Pre-training on TER (large data) and fine-tuning on human scores (small data) is better than training only on human scores (small data).}

Further experiments with limited target-domain data (\Cref{fig:limited_finetuning}) show that the proposed pre-training \& fine-tuning regime does not perform well even with less fine-tuning data.
The same figure also shows fine-tuning sensitivity to the selected data.
Variance is caused by both the optimization process and data subsampling.
Even though we include confidence intervals, in deployment one would start multiple runs and use the best-performing one.
A striking observation is that very little human-annotated data is needed for training.
Further research should more closely examine the relationship between model capacity, data requirements and QE performance.

\begin{table}[htbp]
\centering
\resizebox{\linewidth}{!}
{
\begin{tabular}{lcccc}
\toprule
& \multicolumn{2}{c}{\textbf{ME}} & \multicolumn{2}{c}{\textbf{QE}} \\
\textbf{Metric} & \textbf{Single} & \textbf{Multi} & \textbf{Single} & \textbf{Multi} \\
\midrule
\textbf{sentBLEU} & 60.4\% & 47.0\% & 15.5\% & 23.9\% \\
\textbf{BLEURT} & 59.7\% & \whitezero{}5.7\% & 26.7\% & \whitezero{}5.7\% \\
\textbf{ChrF} & 58.5\% & 42.6\% & 23.7\% & 24.0\% \\
\textbf{METEOR} & 51.6\% & 36.5\% & 22.6\% & 24.5\% \\
\textbf{COMET} & 44.4\% & 23.1\% & 13.1\% & 21.0\% \\
\textbf{TER} & 37.4\% & 36.8\% & \whitezero{}7.6\% & 18.7\% \\
\textbf{human} & - & - & 19.8\% & \whitezero{}5.5\%\\
\bottomrule
\end{tabular}
}
\caption{Pearson correlations between system outputs and automated metrics or human z-scores for either multiple single-metric models (Single) or a single multi-target model (Multi).}
\label{tab:joint_prediction}
\end{table}

\subsection{Joint prediction of multiple metrics}
\label{sec:joint_prediction}

In this section, we investigate using all the automated metrics at the same time in a single model instead of multiple individual models.

For both the WMT News and QE datasets (individually), all the metric scores for a single segment can be predicted at the same time.
Instead of training 6+7 individual models to predict each metric, we train two models (for ME and QE data) that predict all available metrics at once (similar to BLEURT pre-training phase) using different regression heads.
The only difference in the architecture from \Cref{sec:model_details} is that the last linear layer has 6 or 7 output neurons instead of one.
The loss for model $f$ is then defined as $\sum_{m \,\in\, \text{metrics}} \text{MSE}(f_m(s,t), m(s,t,r))$.
Having multiple targets in a single training can provide more signal and better representation \citep{aho2012multi,korneva2020towards}.
The results shown in \Cref{tab:joint_prediction} demonstrate that for the smaller dataset (QE), joint learning mostly helps in metric prediction but not in human z-score prediction.
This may be because of a loss imbalance of 6 target outputs optimizing on automated metrics and only 1 target output optimizing on human z-scores.

\section{Complexity \& Fluency Estimation}
\label{sec:complexity_estimation}

Currently, our model was dependent on mostly the source and the hypothesis.
If it had access to only the hypothesis, it could still consider its fluency and other factors in estimating the metric.
Likewise, having access to only the source would correspond to sentence difficulty/complexity estimation.
Similarly to \citet{wan2022unite,don2022prequel}, we explored both of these modes and found very high sentence-level correlations.\footnote{While performing an NLP task only on part of the input can reveal annotation artifacts \citep{gururangan2018annotation}, our case corresponds more to complexity and fluency estimation.}
\begin{align*}
\rho(\{ (f_\textsc{ME}(h), f_\textsc{sentBLEU}(h, r)) \}) &= 55.5\% \\
\rho(\{ (f_\textsc{ME}(s), f_\textsc{sentBLEU}(h, r)) \}) &= 55.3\%
\end{align*}

These high correlations, which are close to the full text-only model's performance ($\rho=56.5\%$) show that our model is not able to utilize the relationship between the source and the hypothesis and that a more elaborate models should be considered.
These results are in line with general findings of \citet{behnke2022bias}.
The models' inadequacy is also shown by the imperfect performance when given access to the hypothesis and the reference, just as the metric has:
\begin{align*}
\rho(\{ (f_\textsc{ME}(s, h, r), f_\textsc{sentBLEU}(h, r)) \}) &= 60.7\% \\
\rho(\{ (f_\textsc{ME}(h, r), f_\textsc{sentBLEU}(h, r)) \}) &= 61.5\% 
\end{align*}

However, the focus solely on the hypothesis or just the source itself has been confirmed for also other QE systems \citep{sun2020we} and our model is not an outlier.

\section{Related Work}
\label{sec:related_work}

This section discusses how our proposed ME task fits in the field of QE.

\paragraph{Confidence estimation.}
The task of ME has a connection to an older task of confidence estimation \citep{blatz2004confidence}, which predates QE \citep{specia2013quest}.
In confidence estimation, the goal is to predict the probability of the output being correct.
\citet{blatz2004confidence} define correctness as a binary class which is based on two thresholded MT metrics: word error rate and NIST \citep{doddington2002automatic}.
This is in contrast to the ME task which is a regression task (predicting e.g. 0.7 instead of \texttt{GOOD} and segment-level).

More recent works use the term confidence estimation more freely to mean essentially the QE task with full training data and model access \citep{chelba2020data}.
Because this term is used also in other contexts, such as in calibration \citep{wan2020self,wang2020inference}, we define the task {metric estimation} to avoid ambiguity.

\paragraph{Feature-based QE models.}

\citet{specia2010machine,specia2013quest} pioneered the work of mainstream MT QE.
The QuEst model uses support vector regression on top of features such as source \& target lengths, the number of translations in a phrase table or target sentence language model probability.
Further research has been devoted to devising good features for MT QE models, such as grammatical ones \citep{felice2012linguistic}, ones based on the decoder \citep{avramidis2012quality,fomicheva2020unsupervised} or based on the model embeddings spaces \citep{shah2016shef,chen2017improving}. %

\paragraph{Deeper QE models.}

QUETCH \citep{kreutzer2015quality}, NuQE \citep{martins2016unbabel}, DeepQuest \citep{ive2018deepquest} and others \citep{kim2016recurrent,li2018unified} regress directly from the source and hypothesis texts into a score.
Notably some systems approach sentence-level QE by aggregating or otherwise utilizing previously-estimated word-level QE predictions \citep{kepler2019openkiwi}.

\paragraph{Pre-training QE models.}

Closest to our work is computing TER between the hypothesis and its post-edited version \citep{heo2021quality}.
QE is then trained jointly on the artificial and authentic QE data.
Other models are first trained on artificial (pre-training) and then authentic data \citep{baek2020patquest,cui2021directqe,yankovskaya2021direct}.
The pre-training task does not have to be tied to QE.
Large pre-trained language models have also been used for QE \citep{hu2020niutrans,moura2020unbabel,nakamachi2020tmuou,eo2021comparative}.
In contrast, our pre-training aims not only to acquire better sentence representations in general but specifically to acquire better sentence representations for translation quality score estimation.

\paragraph{QE models outside of MT.}
The idea to estimate the quality of a prediction given only itself and the input has also been applied to other NLP tasks.
RUBER \cite{tao2018ruber} uses an RNN-based model to compute a score for a context-response pair in a dialog that is combined with a reference-based score to obtain the final metric.
BERT-RUBER extends this by using pre-trained representations and GRADE uses the unreferenced scorer in combination with a module that constructs a conversational graph using ConceptNet concepts and reasons over it \cite{huang2020grade}.
Recently different works evaluate responses based on specific quality attributes, for example, groundedness \cite{honovich2021q2}, or combine them into one quality score \cite{pang2020holisticeval}.
For open-ended tasks, reference-free metrics, similar to QE, are more desirable because they make fewer assumptions about how the hypothesis should look like.
Compared to those tasks, the admissible hypothesis space in MT for a given source sentence is more constrained.

\section{Discussion} 

The experiments have shown that the outputs of automated metrics can be predicted even without access to the reference.
Pre-training only on {TER} and not any other automated metric outperformed training from scratch, which highlights the importance of exploring multiple metrics instead of just one or two of the most popular ones.
Pre-training on {TER} helped because of the large absolute zero-shot correlation.
We are, however, unable to provide an explanation for this zero-shot correlation in the first place.

Our ME approach can also be potentially used for improving models, even outside of the MT field, by providing additional signals through self-supervision.
For example, the ME model could be used instead of the decoder probabilities for reranking when decoding with beam-search in generative models.
This approach would be similar to using QE in decoding by \citet{fernandes2022quality} but would not require a separate scorer model.

\begin{figure}[htbp]
\includegraphics[width=\linewidth]{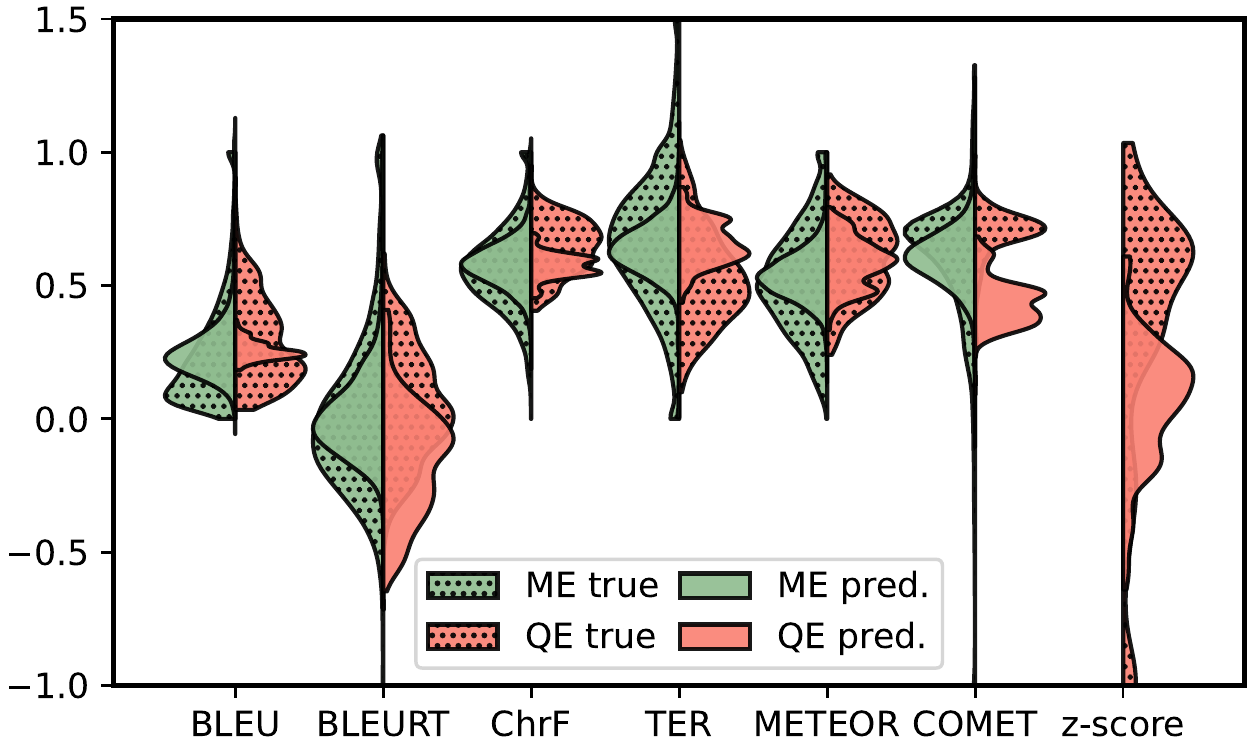}
\caption{Distribution of metric and human judgement values in the used ME and QE datasets together with their predictions. Long tails clipped between -1 and 1.5 for higher resolution.}
\label{fig:metric_violin}
\vspace{-3mm}
\end{figure}

\subsection{Error analysis}
\label{sec:error_analysis}

In this section, we examine three model predictions and comment on their comparison with the target scores.
At the first glance, the model predictions are generally more conservative (lower variance and concentration around the average), as shown in \Cref{fig:metric_violin}.
This is however not an issue when evaluating with Pearson's correlation coefficient as the distributions are rescaled.
While the error analysis serves as a good check for the model outputs, we are unable to clearly define specific failure modes of the model, which does not achieve 100\% metric correlation.

\textexample{
\small
\textbf{Example 1} \hfill  $\,\textbf{sentBLEU}\, 0.08, \textbf{ME}\, 0.07$
\vspace{2mm}

\textbf{Source:} \emph{Police try new, less-lethal tools as protests continue.}

\textbf{Reference:} \emph{Während die Proteste weitergehen, testet die Polizei weniger tödliche Geräte}

\textbf{Reference (lit.):} \emph{While the protest continue, the police is testing less-lethal devices.}

\textbf{Hypothesis:} \emph{Die Polizei probiert neue, weniger tödliche Werkzeuge aus, während die Proteste anhalten.}

\textbf{Hypothesis (lit.):} \emph{The police is testing new, less-lethal tools, as the protest persist.}
}

We first examine cases in which we compare our model's prediction with the true metric value.\footnote{We report true sentBLEU scores and not percentages (i.e. scale is 0-1, not 0-100).}
Example 1 is an almost exact match in sentBLEU and model prediction.
Although the hypothesis is reasonable, it is too literal of a translation.
This is a failure for the automated metric but a success for our model because it predicted the metric accurately.

\textexample{
\small
\textbf{Example 2} \hfill $\,\textbf{sentBLEU}\, 0.67, \textbf{ME}\, -0.51$
\vspace{2mm}

\textbf{Source:} \emph{It didn't work.}

\textbf{Reference:} \emph{Das hat nicht funktioniert.}

\textbf{Reference (lit.):} \emph{That did not function.}

\textbf{Hypothesis:} \emph{IT hat nicht funktioniert.}

\textbf{Hypothesis (lit.):} \emph{THAT/IT did not function.}
}

In some error cases, the model prediction was a better quality estimate than the metric it was trained to estimate.
In Example 2, the word \emph{it} is mistaken for a named entity which distorts the sentence meaning: \emph{it} - pronoun, \emph{IT} - abbreviation \emph{Information Technology} (can also mean the tech support department).

\textexample{
\small
\textbf{Example 3}  \hfill $\,\textbf{Human}\, -4.72, \textbf{QE}\, -0.42$
\vspace{2mm}

\textbf{Source:} \emph{There's mask-shaming and then there's full on assault.}

\textbf{Reference:} \emph{Masken-Shaming ist eine Sache, Körperverletzung eine andere.}

\textbf{Reference (lit.):} \emph{Mask-shaming is one thing, body assault is a different one.}

\textbf{Hypothesis:} \emph{Es gibt Masken-Beschämen und dann gibt es voll auf Angriff.}

\textbf{Hypothesis (lit.):} \emph{There is mask-humiliation and then there is full-on assault.}
}

In certain cases, the model output is very far from human judgement.
In Example 3, the hypothesis contains a phrase \emph{voll auf Angriffe} which seems like a good translation but only at the first sight and is, in fact, incorrect (word-wise translation \emph{full - voll, on - auf, assault - Angriff}).
This may be a reason for the low score by the human annotator and it was not captured at all by our model.

\newcounter{gpuFootnoteId}
\footnotetext{Training one full model takes ${\sim}$3 hours on 500k sentences on NVIDIA GeForce GTX 1080 Ti. Machine translation of the same data takes 120 GPU hours on the same model.}\setcounter{gpuFootnoteId}{\value{footnote}}

\subsection{Negative results}

We also attempted to leverage mBERT representations in the main LSTM-based model via concatenation fusion with the last hidden state, identical to the approach of \citet{zouhar2022fusion} for language modelling.
However, the results ($\rho$ = 60.0\% for sentBLEU) were on par with the main LSTM-based model alone ($\rho$ = 60.4\% for sentBLEU).

We experimented with the expressivity of the used model architecture in \Cref{sec:complexity_estimation} by granting the model access to the source, the reference and the hypothesis.
This should provide sufficient information to be able to learn the specific metrics, however, the model was unable to fit it perfectly.
Approaches in future work should therefore use models with larger capacities and explore specific mechanics that could be useful for predicting automated metrics relying on n-gram overlaps.

\vspace{10mm}

\section{Conclusion}

We proposed the task of \textbf{metric estimation} for machine translation, as a parallel to quality estimation and attempted to solve it with a baseline BiLSTM model.
We show that it is possible to predict the output of a metric without even seeing the reference ($\rho$ = 60.4\% for sentBLEU and $\bar{\rho}$ = 51.3\% for other metrics).
The main advantage of this task compared to QE is that the data for training ME models to predict a particular metric can be generated from any parallel corpus on which the metric can be run.
While pre-training on {TER} outperformed training from scratch, it did not perform better than the commonly used baseline, {COMET-QE}.

\paragraph{Future work.}

Despite the negative results, features in the hypothesis space should be more explored for tasks beyond ME/QE, such as calibration of self-reported confidence in generative models.
Metric and quality estimation could also be a part of the MT system itself (e.g. as a separate head) which would alleviate the need for an external ME/QE model.
The ME models should also be evaluated for cross-domain performance, similarly to our cross-system evaluation as motivated by needs of production settings.
Non-perfect correlation with the metrics when presented with the same input shows the imporants of exploring more complex architectures or optimization approaches.

\section*{Limitations}

Although our model outperformed the simple baselines in ME and QE, it provides less explainability because a specific QE output can not be linked easily to input features.
The model also required much longer training\footnotemark[\value{gpuFootnoteId}] while the baselines just need a simple featurizer, MT intrinsic features and can run the linear regression fitting on a CPU.
Nevertheless, the largest computational bottleneck in this research has been running the MT system inference\footnotemark[\value{gpuFootnoteId}] rather than training the individual models.

Concerns have long been raised about using segment-level metrics/evaluations because of the large variance \citep{lavie2010evaluating}.
However, we find that for automated metrics, our models are still able to deal with this variance.

\section*{Ethics statement}

Detailed error analysis should always be performed before deploying a quality estimation system in machine translation production pipelines.

\section*{Acknowledgements}

We thank Ricardo Rei (\href{https://unbabel.com/}{Unbabel}), Jonas Belouadi (Universität Bielefeld) and Florian Schottmann (\href{https://www.textshuttle.ai/}{TextShuttle}) for their proofreading, discussions and comments on our work.

\bibliography{misc/bibliography.bib}

\begin{thebibliography}{77}
\expandafter\ifx\csname natexlab\endcsname\relax\def\natexlab#1{#1}\fi

\bibitem[{Aho et~al.(2012)Aho, {\v{Z}}enko, D{\v{z}}zeroski, Elomaa, and
  Brodley}]{aho2012multi}
Timo Aho, Bernard {\v{Z}}enko, Sa{\v{s}}o D{\v{z}}zeroski, Tapio Elomaa, and
  Carla Brodley. 2012.
\newblock \href {https://www.jmlr.org/papers/volume13/aho12a/aho12a.pdf}
  {Multi-target regression with rule ensembles.}
\newblock \emph{Journal of Machine Learning Research}, 13(8).

\bibitem[{Avramidis(2012)}]{avramidis2012quality}
Eleftherios Avramidis. 2012.
\newblock \href {https://aclanthology.org/W12-3108} {Quality estimation for
  machine translation output using linguistic analysis and decoding features}.
\newblock In \emph{Proceedings of the seventh workshop on statistical machine
  translation}, pages 84--90.

\bibitem[{Baek et~al.(2020)Baek, Kim, Moon, Kim, and Park}]{baek2020patquest}
Yujin Baek, Zae~Myung Kim, Jihyung Moon, Hyunjoong Kim, and Eunjeong Park.
  2020.
\newblock \href {https://aclanthology.org/2020.wmt-1.113/} {Patquest: {Papago}
  translation quality estimation}.
\newblock In \emph{Proceedings of the Fifth Conference on Machine Translation},
  pages 991--998.

\bibitem[{Banerjee and Lavie(2005)}]{banarjee2005}
Satanjeev Banerjee and Alon Lavie. 2005.
\newblock \href {https://www.aclweb.org/anthology/W05-0909} {{METEOR}: {An}
  automatic metric for {MT} evaluation with improved correlation with human
  judgments}.
\newblock In \emph{Proceedings of the {ACL} Workshop on Intrinsic and Extrinsic
  Evaluation Measures for Machine Translation and/or Summarization}, pages
  65--72, Ann Arbor, Michigan.

\bibitem[{Ba{\~n}{\'o}n et~al.(2020)Ba{\~n}{\'o}n, Chen, Haddow, Heafield,
  Hoang, Espl{\`a}-Gomis, Forcada, Kamran, Kirefu, Koehn, Ortiz~Rojas,
  Pla~Sempere, Ram{\'\i}rez-S{\'a}nchez, Sarr{\'\i}as, Strelec, Thompson,
  Waites, Wiggins, and Zaragoza}]{paracrawl}
Marta Ba{\~n}{\'o}n, Pinzhen Chen, Barry Haddow, Kenneth Heafield, Hieu Hoang,
  Miquel Espl{\`a}-Gomis, Mikel~L. Forcada, Amir Kamran, Faheem Kirefu, Philipp
  Koehn, Sergio Ortiz~Rojas, Leopoldo Pla~Sempere, Gema
  Ram{\'\i}rez-S{\'a}nchez, Elsa Sarr{\'\i}as, Marek Strelec, Brian Thompson,
  William Waites, Dion Wiggins, and Jaume Zaragoza. 2020.
\newblock \href {https://doi.org/10.18653/v1/2020.acl-main.417} {{P}ara{C}rawl:
  {Web}-scale acquisition of parallel corpora}.
\newblock In \emph{Proceedings of the 58th Annual Meeting of the Association
  for Computational Linguistics}, pages 4555--4567, Online.

\bibitem[{Behnke et~al.(2022)Behnke, Fomicheva, and Specia}]{behnke2022bias}
Hanna Behnke, Marina Fomicheva, and Lucia Specia. 2022.
\newblock \href {https://aclanthology.org/2022.acl-long.104/} {Bias mitigation
  in machine translation quality estimation}.
\newblock In \emph{Proceedings of the 60th Annual Meeting of the Association
  for Computational Linguistics (Volume 1: Long Papers)}, pages 1475--1487.

\bibitem[{Blatz et~al.(2004)Blatz, Fitzgerald, Foster, Gandrabur, Goutte,
  Kulesza, Sanchis, and Ueffing}]{blatz2004confidence}
John Blatz, Erin Fitzgerald, George Foster, Simona Gandrabur, Cyril Goutte,
  Alex Kulesza, Alberto Sanchis, and Nicola Ueffing. 2004.
\newblock \href {https://aclanthology.org/C04-1046} {Confidence estimation for
  machine translation}.
\newblock In \emph{Coling 2004: Proceedings of the 20th international
  conference on computational linguistics}, pages 315--321.

\bibitem[{Bojar et~al.(2014)Bojar, Buck, Federmann, Haddow, Koehn, Leveling,
  Monz, Pecina, Post, Saint-Amand, Soricut, Specia, and Tamchyna}]{wmt14}
Ondrej Bojar, Christian Buck, Christian Federmann, Barry Haddow, Philipp Koehn,
  Johannes Leveling, Christof Monz, Pavel Pecina, Matt Post, Herve Saint-Amand,
  Radu Soricut, Lucia Specia, and Ale\v{s} Tamchyna. 2014.
\newblock \href {http://www.aclweb.org/anthology/W/W14/W14-3302} {Findings of
  the 2014 workshop on statistical machine translation}.
\newblock In \emph{Proceedings of the Ninth Workshop on Statistical Machine
  Translation}, pages 12--58, Baltimore, Maryland, USA.

\bibitem[{Chelba et~al.(2020)Chelba, Kazawa, Klingner, Zhou, Niu, and
  Li}]{chelba2020data}
Ciprian Chelba, Hideto Kazawa, Jeff Klingner, Junpei Zhou, Mengmeng Niu, and
  Music Li. 2020.
\newblock \href {https://arxiv.org/abs/2010.13856} {Data troubles in sentence
  level confidence estimation for machine translation}.

\bibitem[{Chen et~al.(2017)Chen, Tan, Zhang, Xiang, Zhang, Li, and
  Wang}]{chen2017improving}
Zhiming Chen, Yiming Tan, Chenlin Zhang, Qingyu Xiang, Lilin Zhang, Maoxi Li,
  and Mingwen Wang. 2017.
\newblock \href {https://aclanthology.org/W17-4761} {Improving machine
  translation quality estimation with neural network features}.
\newblock In \emph{Proceedings of the Second Conference on Machine
  Translation}, pages 551--555.

\bibitem[{Cui et~al.(2021)Cui, Huang, Li, Geng, Zheng, Huang, and
  Chen}]{cui2021directqe}
Qu~Cui, Shujian Huang, Jiahuan Li, Xiang Geng, Zaixiang Zheng, Guoping Huang,
  and Jiajun Chen. 2021.
\newblock \href {https://ojs.aaai.org/index.php/AAAI/article/view/17506}
  {Directqe: {Direct} pretraining for machine translation quality estimation}.
\newblock In \emph{Proceedings of the AAAI Conference on Artificial
  Intelligence}, volume~35, pages 12719--12727.

\bibitem[{de~Souza et~al.(2014)de~Souza, Turchi, and Negri}]{de2014machine}
Jos{\'e}~GC de~Souza, Marco Turchi, and Matteo Negri. 2014.
\newblock \href {https://aclanthology.org/C14-1040/} {Machine translation
  quality estimation across domains}.
\newblock In \emph{Proceedings of COLING 2014, the 25th International
  Conference on Computational Linguistics: Technical Papers}, pages 409--420.

\bibitem[{Devlin et~al.(2019)Devlin, Chang, Lee, and
  Toutanova}]{devlin2019bert}
Jacob Devlin, Ming-Wei Chang, Kenton Lee, and Kristina Toutanova. 2019.
\newblock \href {https://doi.org/10.18653/v1/N19-1423} {{BERT}: {Pre}-training
  of deep bidirectional transformers for language understanding}.
\newblock In \emph{Proceedings of the 2019 Conference of the North {A}merican
  Chapter of the Association for Computational Linguistics: Human Language
  Technologies, Volume 1 (Long and Short Papers)}, pages 4171--4186,
  Minneapolis, Minnesota.

\bibitem[{Doddington(2002)}]{doddington2002automatic}
George Doddington. 2002.
\newblock \href {https://dl.acm.org/doi/abs/10.5555/1289189.1289273} {Automatic
  evaluation of machine translation quality using n-gram co-occurrence
  statistics}.
\newblock In \emph{Proceedings of the second international conference on Human
  Language Technology Research}, pages 138--145.

\bibitem[{Don-Yehiya et~al.(2022)Don-Yehiya, Choshen, and
  Abend}]{don2022prequel}
Shachar Don-Yehiya, Leshem Choshen, and Omri Abend. 2022.
\newblock \href {https://arxiv.org/abs/2205.09178/} {Prequel: Quality
  estimation of machine translation outputs in advance}.
\newblock \emph{arXiv preprint arXiv:2205.09178}.

\bibitem[{Eo et~al.(2021)Eo, Park, Moon, Seo, and Lim}]{eo2021comparative}
Sugyeong Eo, Chanjun Park, Hyeonseok Moon, Jaehyung Seo, and Heuiseok Lim.
  2021.
\newblock \href {https://search.ieice.org/bin/summary.php?id=e101-d_9_2417}
  {Comparative analysis of current approaches to quality estimation for neural
  machine translation}.
\newblock \emph{Applied Sciences}, 11(14):6584.

\bibitem[{Fan et~al.(2021)Fan, Bhosale, Schwenk, Ma, El-Kishky, Goyal, Baines,
  Celebi, Wenzek, Chaudhary et~al.}]{fan2021beyond}
Angela Fan, Shruti Bhosale, Holger Schwenk, Zhiyi Ma, Ahmed El-Kishky,
  Siddharth Goyal, Mandeep Baines, Onur Celebi, Guillaume Wenzek, Vishrav
  Chaudhary, et~al. 2021.
\newblock \href {https://www.jmlr.org/papers/volume22/20-1307/20-1307.pdf}
  {Beyond english-centric multilingual machine translation.}
\newblock \emph{J. Mach. Learn. Res.}, 22(107):1--48.

\bibitem[{Felice and Specia(2012)}]{felice2012linguistic}
Mariano Felice and Lucia Specia. 2012.
\newblock \href {https://aclanthology.org/W12-3110} {Linguistic features for
  quality estimation}.
\newblock In \emph{Proceedings of the Seventh Workshop on Statistical Machine
  Translation}, pages 96--103.

\bibitem[{Fernandes et~al.(2022)Fernandes, Farinhas, Rei, de~Souza, Ogayo,
  Neubig, and Martins}]{fernandes2022quality}
Patrick Fernandes, Ant{\'o}nio Farinhas, Ricardo Rei, Jos{\'e}~GC de~Souza,
  Perez Ogayo, Graham Neubig, and Andr{\'e}~FT Martins. 2022.
\newblock \href {https://arxiv.org/abs/2205.00978} {Quality-aware decoding for
  neural machine translation}.
\newblock \emph{arXiv preprint arXiv:2205.00978}.

\bibitem[{Fomicheva et~al.(2020)Fomicheva, Sun, Yankovskaya, Blain, Guzm{\'a}n,
  Fishel, Aletras, Chaudhary, and Specia}]{fomicheva2020unsupervised}
Marina Fomicheva, Shuo Sun, Lisa Yankovskaya, Fr{\'e}d{\'e}ric Blain, Francisco
  Guzm{\'a}n, Mark Fishel, Nikolaos Aletras, Vishrav Chaudhary, and Lucia
  Specia. 2020.
\newblock \href {https://aclanthology.org/2020.tacl-1.35/} {Unsupervised
  quality estimation for neural machine translation}.
\newblock \emph{Transactions of the Association for Computational Linguistics},
  8:539--555.

\bibitem[{Gage(1994)}]{gage1994new}
Philip Gage. 1994.
\newblock \href {https://www.derczynski.com/papers/archive/BPE_Gage.pdf} {A new
  algorithm for data compression}.
\newblock \emph{C Users Journal}, 12(2):23--38.

\bibitem[{Gehring et~al.(2017)Gehring, Auli, Grangier, Yarats, and
  Dauphin}]{gehring2017convolutional}
Jonas Gehring, Michael Auli, David Grangier, Denis Yarats, and Yann~N Dauphin.
  2017.
\newblock \href {http://proceedings.mlr.press/v70/gehring17a.html}
  {Convolutional sequence to sequence learning}.
\newblock In \emph{Proceedings of the 34th International Conference on Machine
  Learning-Volume 70}, pages 1243--1252.

\bibitem[{Glushkova et~al.(2021)Glushkova, Zerva, Rei, and
  Martins}]{glushkova2021uncertainty}
Taisiya Glushkova, Chrysoula Zerva, Ricardo Rei, and Andr{\'e}~FT Martins.
  2021.
\newblock \href {https://aclanthology.org/2021.findings-emnlp.330/}
  {Uncertainty-aware machine translation evaluation}.
\newblock In \emph{Findings of the Association for Computational Linguistics:
  EMNLP 2021}, pages 3920--3938.

\bibitem[{Gururangan et~al.(2018)Gururangan, Swayamdipta, Levy, Schwartz,
  Bowman, and Smith}]{gururangan2018annotation}
Suchin Gururangan, Swabha Swayamdipta, Omer Levy, Roy Schwartz, Samuel Bowman,
  and Noah~A Smith. 2018.
\newblock \href {https://aclanthology.org/N18-2017/} {Annotation artifacts in
  natural language inference data}.
\newblock In \emph{Proceedings of the 2018 Conference of the North American
  Chapter of the Association for Computational Linguistics: Human Language
  Technologies, Volume 2 (Short Papers)}, pages 107--112.

\bibitem[{Heo et~al.(2021)Heo, Lee, Jung, and Lee}]{heo2021quality}
Dam Heo, WonKee Lee, Baikjin Jung, and Jong-Hyeok Lee. 2021.
\newblock \href {https://aclanthology.org/2021.wmt-1.96/} {Quality estimation
  using dual encoders with transfer learning}.
\newblock In \emph{Proceedings of the Sixth Conference on Machine Translation},
  pages 920--927.

\bibitem[{Honovich et~al.(2021)Honovich, Choshen, Aharoni, Neeman, Szpektor,
  and Abend}]{honovich2021q2}
Or~Honovich, Leshem Choshen, Roee Aharoni, Ella Neeman, Idan Szpektor, and Omri
  Abend. 2021.
\newblock \href {https://doi.org/10.18653/v1/2021.emnlp-main.619} {$q^{2}$:
  {E}valuating factual consistency in knowledge-grounded dialogues via question
  generation and question answering}.
\newblock In \emph{Proceedings of the 2021 Conference on Empirical Methods in
  Natural Language Processing}, pages 7856--7870, Online and Punta Cana,
  Dominican Republic.

\bibitem[{Hu et~al.(2020)Hu, Liu, Feng, Xu, Xu, Zhou, Yan, Luo, Wang, Meng
  et~al.}]{hu2020niutrans}
Chi Hu, Hui Liu, Kai Feng, Chen Xu, Nuo Xu, Zefan Zhou, Shiqin Yan, Yingfeng
  Luo, Chenglong Wang, Xia Meng, et~al. 2020.
\newblock \href {https://aclanthology.org/2020.wmt-1.117/} {The niutrans system
  for the {WMT20} quality estimation shared task}.
\newblock In \emph{Proceedings of the Fifth Conference on Machine Translation},
  pages 1018--1023.

\bibitem[{Huang et~al.(2020)Huang, Ye, Qin, Lin, and Liang}]{huang2020grade}
Lishan Huang, Zheng Ye, Jinghui Qin, Liang Lin, and Xiaodan Liang. 2020.
\newblock \href {https://aclanthology.org/2020.emnlp-main.742/} {{GRADE}:
  {Automatic} graph-enhanced coherence metric for evaluating open-domain
  dialogue systems}.
\newblock \emph{arXiv preprint arXiv:2010.03994}.

\bibitem[{Ive et~al.(2018)Ive, Blain, and Specia}]{ive2018deepquest}
Julia Ive, Fr{\'e}d{\'e}ric Blain, and Lucia Specia. 2018.
\newblock \href {https://aclanthology.org/C18-1266/} {{DeepQuest}: {A}
  framework for neural-based quality estimation}.
\newblock In \emph{Proceedings of the 27th International Conference on
  Computational Linguistics}, pages 3146--3157.

\bibitem[{Kepler et~al.(2019)Kepler, Tr{\'e}nous, Treviso, Vera, and
  Martins}]{kepler2019openkiwi}
F{\'a}bio Kepler, Jonay Tr{\'e}nous, Marcos Treviso, Miguel Vera, and
  Andr{\'e}~FT Martins. 2019.
\newblock \href {https://aclanthology.org/P19-3.pdf#page=131} {{OpenKiwi}: {An}
  open source framework for quality estimation}.
\newblock \emph{ACL 2019}, page 117.

\bibitem[{Kim and Lee(2016)}]{kim2016recurrent}
Hyun Kim and Jong-Hyeok Lee. 2016.
\newblock \href {https://aclanthology.org/W16-2384/} {Recurrent neural network
  based translation quality estimation}.
\newblock In \emph{Proceedings of the First Conference on Machine Translation:
  Volume 2, Shared Task Papers}, pages 787--792.

\bibitem[{Kingma and Ba(2015)}]{Kingma2015AdamAM}
Diederik~P. Kingma and Jimmy Ba. 2015.
\newblock \href {https://arxiv.org/abs/1412.6980} {Adam: {A} method for
  stochastic optimization}.
\newblock \emph{CoRR}, abs/1412.6980.

\bibitem[{Kocmi et~al.(2021)Kocmi, Federmann, Grundkiewicz, Junczys-Dowmunt,
  Matsushita, and Menezes}]{kocmi2021ship}
Tom Kocmi, Christian Federmann, Roman Grundkiewicz, Marcin Junczys-Dowmunt,
  Hitokazu Matsushita, and Arul Menezes. 2021.
\newblock \href {https://aclanthology.org/2021.wmt-1.57/} {To ship or not to
  ship: {An} extensive evaluation of automatic metrics for machine
  translation}.
\newblock In \emph{Proceedings of the Sixth Conference on Machine Translation},
  pages 478--494.

\bibitem[{Korneva and Blockeel(2020)}]{korneva2020towards}
Evgeniya Korneva and Hendrik Blockeel. 2020.
\newblock \href
  {https://link.springer.com/chapter/10.1007/978-3-030-65965-3_23} {Towards
  better evaluation of multi-target regression models}.
\newblock In \emph{Joint European Conference on Machine Learning and Knowledge
  Discovery in Databases}, pages 353--362. Springer.

\bibitem[{Kreutzer et~al.(2015)Kreutzer, Schamoni, and
  Riezler}]{kreutzer2015quality}
Julia Kreutzer, Shigehiko Schamoni, and Stefan Riezler. 2015.
\newblock \href {https://aclanthology.org/W15-3037/} {Quality estimation from
  scratch (quetch): {Deep} learning for word-level translation quality
  estimation}.
\newblock In \emph{Proceedings of the Tenth Workshop on Statistical Machine
  Translation}, pages 316--322.

\bibitem[{Lavie(2010)}]{lavie2010evaluating}
Alon Lavie. 2010.
\newblock \href {https://aclanthology.org/2010.amta-tutorials.4/} {Evaluating
  the output of machine translation systems}.
\newblock In \emph{Proceedings of the 9th Conference of the Association for
  Machine Translation in the Americas: Tutorials}.

\bibitem[{Li et~al.(2018)Li, Xiang, Chen, and Wang}]{li2018unified}
Maoxi Li, Qingyu Xiang, Zhiming Chen, and Mingwen Wang. 2018.
\newblock \href {https://search.ieice.org/bin/summary.php?id=e101-d_9_2417} {A
  unified neural network for quality estimation of machine translation}.
\newblock \emph{IEICE TRANSACTIONS on Information and Systems},
  101(9):2417--2421.

\bibitem[{Ma et~al.(2019)Ma, Wei, Bojar, and Graham}]{ma2019results}
Qingsong Ma, Johnny Wei, Ond{\v{r}}ej Bojar, and Yvette Graham. 2019.
\newblock \href {https://doras.dcu.ie/24262/1/WMT02.pdf} {Results of the
  {WMT19} metrics shared task: {Segment}-level and strong {MT} systems pose big
  challenges}.
\newblock In \emph{Proceedings of the Fourth Conference on Machine Translation
  (Volume 2: Shared Task Papers, Day 1)}, pages 62--90.

\bibitem[{Martins et~al.(2016)Martins, Astudillo, Hokamp, and
  Kepler}]{martins2016unbabel}
Andr{\'e}~FT Martins, Ram{\'o}n~Fernandez Astudillo, Chris Hokamp, and Fabio
  Kepler. 2016.
\newblock \href {https://aclanthology.org/W16-2387} {Unbabel’s participation
  in the wmt16 word-level translation quality estimation shared task}.
\newblock In \emph{Proceedings of the First Conference on Machine Translation:
  Volume 2, Shared Task Papers}, pages 806--811.

\bibitem[{Mathur et~al.(2020)Mathur, Wei, Freitag, Ma, and
  Bojar}]{mathur2020results}
Nitika Mathur, Johnny Wei, Markus Freitag, Qingsong Ma, and Ond{\v{r}}ej Bojar.
  2020.
\newblock \href {https://aclanthology.org/2020.wmt-1.77/} {Results of the
  {WMT20} metrics shared task}.
\newblock In \emph{Proceedings of the Fifth Conference on Machine Translation},
  pages 688--725.

\bibitem[{Moura et~al.(2020)Moura, Vera, van Stigt, Kepler, and
  Martins}]{moura2020unbabel}
Joao Moura, Miguel Vera, Daan van Stigt, Fabio Kepler, and Andr{\'e}~FT
  Martins. 2020.
\newblock \href {https://aclanthology.org/2020.wmt-1.119/} {Ist-unbabel
  participation in the wmt20 quality estimation shared task}.
\newblock In \emph{Proceedings of the Fifth Conference on Machine Translation},
  pages 1029--1036.

\bibitem[{Murgolo et~al.(2022)Murgolo, Sharami, and
  Shterionov}]{murgolo2022quality}
Elena Murgolo, Javad Pourmostafa~Roshan Sharami, and Dimitar Shterionov. 2022.
\newblock \href {https://aclanthology.org/2022.eamt-1.43/} {A quality
  estimation and quality evaluation tool for the translation industry}.
\newblock In \emph{Proceedings of the 23rd Annual Conference of the European
  Association for Machine Translation}, pages 305--306.

\bibitem[{Nakamachi et~al.(2020)Nakamachi, Shimanaka, Kajiwara, and
  Komachi}]{nakamachi2020tmuou}
Akifumi Nakamachi, Hiroki Shimanaka, Tomoyuki Kajiwara, and Mamoru Komachi.
  2020.
\newblock \href {https://aclanthology.org/2020.wmt-1.120/} {Tmuou submission
  for wmt20 quality estimation shared task}.
\newblock In \emph{Proceedings of the Fifth Conference on Machine Translation},
  pages 1037--1041.

\bibitem[{Ng et~al.(2019)Ng, Yee, Baevski, Ott, Auli, and
  Edunov}]{ng2019facebook}
Nathan Ng, Kyra Yee, Alexei Baevski, Myle Ott, Michael Auli, and Sergey Edunov.
  2019.
\newblock \href {https://aclanthology.org/W19-5333/} {Facebook {FAIR}’s
  {WMT19} news translation task submission}.
\newblock In \emph{Proceedings of the Fourth Conference on Machine Translation
  (Volume 2: Shared Task Papers, Day 1)}, pages 314--319.

\bibitem[{Ott et~al.(2019)Ott, Edunov, Baevski, Fan, Gross, Ng, Grangier, and
  Auli}]{ott2019fairseq}
Myle Ott, Sergey Edunov, Alexei Baevski, Angela Fan, Sam Gross, Nathan Ng,
  David Grangier, and Michael Auli. 2019.
\newblock \href {https://aclanthology.org/N19-4009/} {fairseq: {A} fast,
  extensible toolkit for sequence modeling}.
\newblock In \emph{Proceedings of the 2019 Conference of the North American
  Chapter of the Association for Computational Linguistics (Demonstrations)},
  pages 48--53.

\bibitem[{Ott et~al.(2018)Ott, Edunov, Grangier, and Auli}]{ott2018scaling}
Myle Ott, Sergey Edunov, David Grangier, and Michael Auli. 2018.
\newblock \href {https://doi.org/10.18653/v1/W18-6301} {Scaling neural machine
  translation}.
\newblock In \emph{Proceedings of the Third Conference on Machine Translation:
  Research Papers}, pages 1--9, Brussels, Belgium.

\bibitem[{Pang et~al.(2020)Pang, Nijkamp, Han, Zhou, Liu, and
  Tu}]{pang2020holisticeval}
Bo~Pang, Erik Nijkamp, Wenjuan Han, Linqi Zhou, Yixian Liu, and Kewei Tu. 2020.
\newblock \href {https://doi.org/10.18653/v1/2020.acl-main.333} {Towards
  holistic and automatic evaluation of open-domain dialogue generation}.
\newblock In \emph{Proceedings of the 58th Annual Meeting of the Association
  for Computational Linguistics}, pages 3619--3629, Online.

\bibitem[{Papineni et~al.(2002)Papineni, Roukos, Ward, and
  Zhu}]{papineni2002bleu}
Kishore Papineni, Salim Roukos, Todd Ward, and Wei-Jing Zhu. 2002.
\newblock \href {https://aclanthology.org/P02-1040} {{BLEU}: {A} method for
  automatic evaluation of machine translation}.
\newblock In \emph{Proceedings of the 40th annual meeting of the Association
  for Computational Linguistics}, pages 311--318.

\bibitem[{Popovi{\'c}(2015)}]{popovic2015chrf}
Maja Popovi{\'c}. 2015.
\newblock \href {https://aclanthology.org/W15-3049} {{chrF}: {character} n-gram
  f-score for automatic {MT} evaluation}.
\newblock In \emph{Proceedings of the Tenth Workshop on Statistical Machine
  Translation}, pages 392--395.

\bibitem[{Post(2018)}]{post2018call}
Matt Post. 2018.
\newblock \href {https://aclanthology.org/W18-6319/} {A call for clarity in
  reporting {BLEU} scores}.
\newblock In \emph{Proceedings of the Third Conference on Machine Translation:
  Research Papers}, pages 186--191.

\bibitem[{Raffel et~al.(2020)Raffel, Shazeer, Roberts, Lee, Narang, Matena,
  Zhou, Li, Liu et~al.}]{raffel2020exploring}
Colin Raffel, Noam Shazeer, Adam Roberts, Katherine Lee, Sharan Narang, Michael
  Matena, Yanqi Zhou, Wei Li, Peter~J Liu, et~al. 2020.
\newblock \href {https://www.jmlr.org/papers/volume21/20-074/20-074.pdf}
  {Exploring the limits of transfer learning with a unified text-to-text
  transformer.}
\newblock \emph{J. Mach. Learn. Res.}, 21(140):1--67.

\bibitem[{Rei et~al.(2020{\natexlab{a}})Rei, Stewart, Farinha, and
  Lavie}]{rei2020comet}
Ricardo Rei, Craig Stewart, Ana~C Farinha, and Alon Lavie. 2020{\natexlab{a}}.
\newblock \href {https://aclanthology.org/2020.emnlp-main.213/} {{COMET}: {A}
  neural framework for {MT} evaluation}.
\newblock In \emph{Proceedings of the 2020 Conference on Empirical Methods in
  Natural Language Processing (EMNLP)}, pages 2685--2702.

\bibitem[{Rei et~al.(2020{\natexlab{b}})Rei, Stewart, Farinha, and
  Lavie}]{rei-EtAl:2020:WMT}
Ricardo Rei, Craig Stewart, Ana~C Farinha, and Alon Lavie. 2020{\natexlab{b}}.
\newblock \href {https://aclanthology.org/2020.wmt-1.101/} {Unbabel's
  participation in the {WMT20} metrics shared task}.
\newblock In \emph{Proceedings of the Fifth Conference on Machine Translation},
  pages 909--918, Online.

\bibitem[{Rubino and Sumita(2020)}]{rubino2020intermediate}
Raphael Rubino and Eiichiro Sumita. 2020.
\newblock \href {https://aclanthology.org/2020.coling-main.385/} {Intermediate
  self-supervised learning for machine translation quality estimation}.
\newblock In \emph{Proceedings of the 28th International Conference on
  Computational Linguistics}, pages 4355--4360.

\bibitem[{Schwenk et~al.(2021)Schwenk, Wenzek, Edunov, Grave, Joulin, and
  Fan}]{schwenk2021ccmatrix}
Holger Schwenk, Guillaume Wenzek, Sergey Edunov, {\'E}douard Grave, Armand
  Joulin, and Angela Fan. 2021.
\newblock \href {https://aclanthology.org/2021.acl-long.507/} {{CCMatrix}:
  {Mining} billions of high-quality parallel sentences on the web}.
\newblock In \emph{Proceedings of the 59th Annual Meeting of the Association
  for Computational Linguistics and the 11th International Joint Conference on
  Natural Language Processing (Volume 1: Long Papers)}, pages 6490--6500.

\bibitem[{Sellam et~al.(2020)Sellam, Das, and Parikh}]{sellam2020bleurt}
Thibault Sellam, Dipanjan Das, and Ankur Parikh. 2020.
\newblock \href {https://aclanthology.org/2020.acl-main.704} {{BLEURT}:
  {Learning} robust metrics for text generation}.
\newblock In \emph{Proceedings of the 58th Annual Meeting of the Association
  for Computational Linguistics}, pages 7881--7892.

\bibitem[{Sennrich et~al.(2016)Sennrich, Haddow, and
  Birch}]{Sennrich2016NeuralMT}
Rico Sennrich, Barry Haddow, and Alexandra Birch. 2016.
\newblock \href {https://aclanthology.org/P16-1162/} {Neural machine
  translation of rare words with subword units}.
\newblock \emph{ArXiv}, abs/1508.07909.

\bibitem[{Shah et~al.(2016)Shah, Bougares, Barrault, and Specia}]{shah2016shef}
Kashif Shah, Fethi Bougares, Lo{\"\i}c Barrault, and Lucia Specia. 2016.
\newblock \href {https://aclanthology.org/W16-2392} {Shef-lium-nn: {Sentence}
  level quality estimation with neural network features}.
\newblock In \emph{Proceedings of the First Conference on Machine Translation:
  Volume 2, Shared Task Papers}, pages 838--842.

\bibitem[{Snover et~al.(2006)Snover, Dorr, Schwartz, Micciulla, and
  Makhoul}]{snover2006study}
Matthew Snover, Bonnie Dorr, Richard Schwartz, Linnea Micciulla, and John
  Makhoul. 2006.
\newblock \href {https://aclanthology.org/2006.amta-papers.25/} {A study of
  translation edit rate with targeted human annotation}.
\newblock In \emph{Proceedings of the 7th Conference of the Association for
  Machine Translation in the Americas: Technical Papers}, pages 223--231.

\bibitem[{Specia et~al.(2020)Specia, Blain, Fomicheva, Fonseca, Chaudhary,
  Guzm{\'a}n, and Martins}]{specia2020findings}
Lucia Specia, Fr{\'e}d{\'e}ric Blain, Marina Fomicheva, Erick Fonseca, Vishrav
  Chaudhary, Francisco Guzm{\'a}n, and Andr{\'e} F.~T. Martins. 2020.
\newblock \href {https://aclanthology.org/2020.wmt-1.79} {Findings of the {WMT}
  2020 shared task on quality estimation}.
\newblock In \emph{Proceedings of the Fifth Conference on Machine Translation},
  pages 743--764, Online.

\bibitem[{Specia et~al.(2021)Specia, Blain, Fomicheva, Zerva, Li, Chaudhary,
  and Martins}]{specia2021findings}
Lucia Specia, Fr{\'e}d{\'e}ric Blain, Marina Fomicheva, Chrysoula Zerva,
  Zhenhao Li, Vishrav Chaudhary, and Andr{\'e}~FT Martins. 2021.
\newblock \href {https://aclanthology.org/2021.wmt-1.71/} {Findings of the
  {WMT} 2021 shared task on quality estimation}.
\newblock In \emph{Proceedings of the Sixth Conference on Machine Translation},
  pages 684--725.

\bibitem[{Specia et~al.(2010)Specia, Raj, and Turchi}]{specia2010machine}
Lucia Specia, Dhwaj Raj, and Marco Turchi. 2010.
\newblock \href {https://link.springer.com/article/10.1007/s10590-010-9077-2}
  {Machine translation evaluation versus quality estimation}.
\newblock \emph{Machine translation}, 24(1):39--50.

\bibitem[{Specia et~al.(2018)Specia, Scarton, and Paetzold}]{specia2018quality}
Lucia Specia, Carolina Scarton, and Gustavo~Henrique Paetzold. 2018.
\newblock \href
  {https://www.morganclaypool.com/doi/abs/10.2200/S00854ED1V01Y201805HLT039}
  {Quality estimation for machine translation}.
\newblock \emph{Synthesis Lectures on Human Language Technologies},
  11(1):1--162.

\bibitem[{Specia et~al.(2013)Specia, Shah, De~Souza, and
  Cohn}]{specia2013quest}
Lucia Specia, Kashif Shah, Jos{\'e}~GC De~Souza, and Trevor Cohn. 2013.
\newblock \href {https://aclanthology.org/P13-4014} {{QuEst}: {A} translation
  quality estimation framework}.
\newblock In \emph{Proceedings of the 51st Annual Meeting of the Association
  for Computational Linguistics: System Demonstrations}, pages 79--84.

\bibitem[{Sun et~al.(2020)Sun, Guzm{\'a}n, and Specia}]{sun2020we}
Shuo Sun, Francisco Guzm{\'a}n, and Lucia Specia. 2020.
\newblock \href {https://aclanthology.org/2020.acl-main.558/} {Are we
  estimating or guesstimating translation quality?}
\newblock In \emph{Proceedings of the 58th annual meeting of the association
  for computational linguistics}, pages 6262--6267.

\bibitem[{Tamchyna(2021)}]{tamchyna2021deploying}
Ale{\v{s}} Tamchyna. 2021.
\newblock \href
  {https://aclanthology.org/attachments/2021.mtsummit-up.21.Presentation.pdf}
  {Deploying {MT} quality estimation on a large scale: {Lessons} learned and
  open questions}.
\newblock In \emph{Proceedings of Machine Translation Summit XVIII: Users and
  Providers Track}, pages 291--305.

\bibitem[{Tao et~al.(2018)Tao, Mou, Zhao, and Yan}]{tao2018ruber}
Chongyang Tao, Lili Mou, Dongyan Zhao, and Rui Yan. 2018.
\newblock \href
  {https://www.aaai.org/ocs/index.php/AAAI/AAAI18/paper/viewFile/16179/15752}
  {Ruber: {An} unsupervised method for automatic evaluation of open-domain
  dialog systems}.
\newblock In \emph{Thirty-Second AAAI Conference on Artificial Intelligence}.

\bibitem[{Tiedemann(2012)}]{opus_paracrawl}
Jörg Tiedemann. 2012.
\newblock \href
  {http://www.lrec-conf.org/proceedings/lrec2012/pdf/463_Paper.pdf} {Parallel
  data, tools and interfaces in {OPUS}}.
\newblock In \emph{Proceedings of the Eight International Conference on
  Language Resources and Evaluation (LREC'12)}, Istanbul, Turkey.

\bibitem[{Wan et~al.(2022)Wan, Liu, Yang, Zhang, Chen, Wong, and
  Chao}]{wan2022unite}
Yu~Wan, Dayiheng Liu, Baosong Yang, Haibo Zhang, Boxing Chen, Derek Wong, and
  Lidia Chao. 2022.
\newblock \href {https://arxiv.org/abs/2204.13346} {{UniTE}: {Unified}
  translation evaluation}.
\newblock In \emph{Proceedings of the 60th Annual Meeting of the Association
  for Computational Linguistics (Volume 1: Long Papers)}, pages 8117--8127.

\bibitem[{Wan et~al.(2020)Wan, Yang, Wong, Zhou, Chao, Zhang, and
  Chen}]{wan2020self}
Yu~Wan, Baosong Yang, Derek~F Wong, Yikai Zhou, Lidia~S Chao, Haibo Zhang, and
  Boxing Chen. 2020.
\newblock \href {https://aclanthology.org/2020.emnlp-main.80/} {Self-paced
  learning for neural machine translation}.
\newblock In \emph{Proceedings of the 2020 Conference on Empirical Methods in
  Natural Language Processing (EMNLP)}, pages 1074--1080.

\bibitem[{Wang et~al.(2020)Wang, Tu, Shi, and Liu}]{wang2020inference}
Shuo Wang, Zhaopeng Tu, Shuming Shi, and Yang Liu. 2020.
\newblock \href {https://aclanthology.org/2020.acl-main.278/} {On the inference
  calibration of neural machine translation}.
\newblock In \emph{Proceedings of the 58th Annual Meeting of the Association
  for Computational Linguistics}, pages 3070--3079.

\bibitem[{Wu et~al.(2019)Wu, Fan, Baevski, Dauphin, and Auli}]{wu2019pay}
Felix Wu, Angela Fan, Alexei Baevski, Yann~N Dauphin, and Michael Auli. 2019.
\newblock \href {https://arxiv.org/abs/1901.10430} {Pay less attention with
  lightweight and dynamic convolutions}.
\newblock \emph{arXiv preprint arXiv:1901.10430}.

\bibitem[{Xu et~al.(2011)Xu, Povey, Mangu, and Zhu}]{xu2011minimum}
Haihua Xu, Daniel Povey, Lidia Mangu, and Jie Zhu. 2011.
\newblock \href
  {https://www.sciencedirect.com/science/article/pii/S0885230811000192}
  {Minimum bayes risk decoding and system combination based on a recursion for
  edit distance}.
\newblock \emph{Computer Speech \& Language}, 25(4):802--828.

\bibitem[{Yankovskaya et~al.(2018)Yankovskaya, T{\"a}ttar, and
  Fishel}]{yankovskaya2018quality}
Elizaveta Yankovskaya, Andre T{\"a}ttar, and Mark Fishel. 2018.
\newblock \href {https://aclanthology.org/W18-6466/} {Quality estimation with
  force-decoded attention and cross-lingual embeddings}.
\newblock In \emph{Proceedings of the Third Conference on Machine Translation:
  Shared Task Papers}, pages 816--821.

\bibitem[{Yankovskaya and Fishel(2021)}]{yankovskaya2021direct}
Lisa Yankovskaya and Mark Fishel. 2021.
\newblock \href {https://aclanthology.org/2021.wmt-1.101/} {Direct exploitation
  of attention weights for translation quality estimation}.
\newblock In \emph{Proceedings of the Sixth Conference on Machine Translation},
  pages 955--960.

\bibitem[{Zouhar et~al.(2021)Zouhar, Nov{\'a}k, {\v{Z}}ilinec, Bojar,
  Obreg{\'o}n, Hill, Blain, Fomicheva, Specia, and
  Yankovskaya}]{zouhar2021backtranslation}
Vil{\'e}m Zouhar, Michal Nov{\'a}k, Mat{\'u}{\v{s}} {\v{Z}}ilinec, Ond{\v{r}}ej
  Bojar, Mateo Obreg{\'o}n, Robin~L Hill, Fr{\'e}d{\'e}ric Blain, Marina
  Fomicheva, Lucia Specia, and Lisa Yankovskaya. 2021.
\newblock \href {https://aclanthology.org/2021.naacl-main.14/} {Backtranslation
  feedback improves user confidence in {MT}, not quality}.
\newblock In \emph{Proceedings of the 2021 Conference of the North American
  Chapter of the Association for Computational Linguistics: Human Language
  Technologies}, pages 151--161.

\bibitem[{Zouhar et~al.(2022)Zouhar, Mosbach, and Klakow}]{zouhar2022fusion}
Vilém Zouhar, Marius Mosbach, and Dietrich Klakow. 2022.
\newblock \href {https://doi.org/10.48550/ARXIV.2208.02402} {Fusing sentence
  embeddings into {LSTM}-based autoregressive language models}.

\end{thebibliography}
\bibliographystyle{misc/acl_natbib}

\clearpage

\appendix

\begin{figure}[htbp]
\centering
\includegraphics[width=0.7\linewidth]{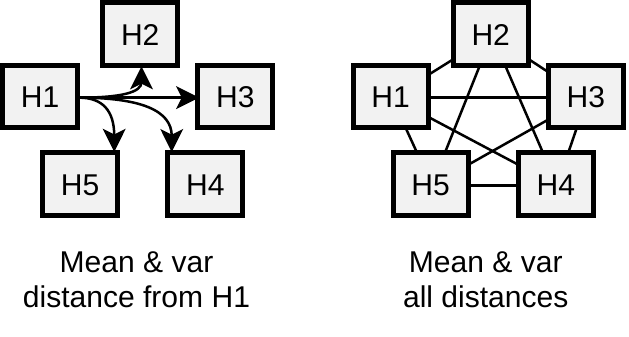}
\caption{Four features based on properties of the generated hypothesis space: mean and variance of either H1 (top hypothesis) or all pairs. The used metric (any hyp-ref-based) considers one of the hypotheses in a pair as the reference.}
\label{fig:features_hyp_space}
\end{figure}

\begin{figure}[htbp]
\includegraphics[width=\linewidth]{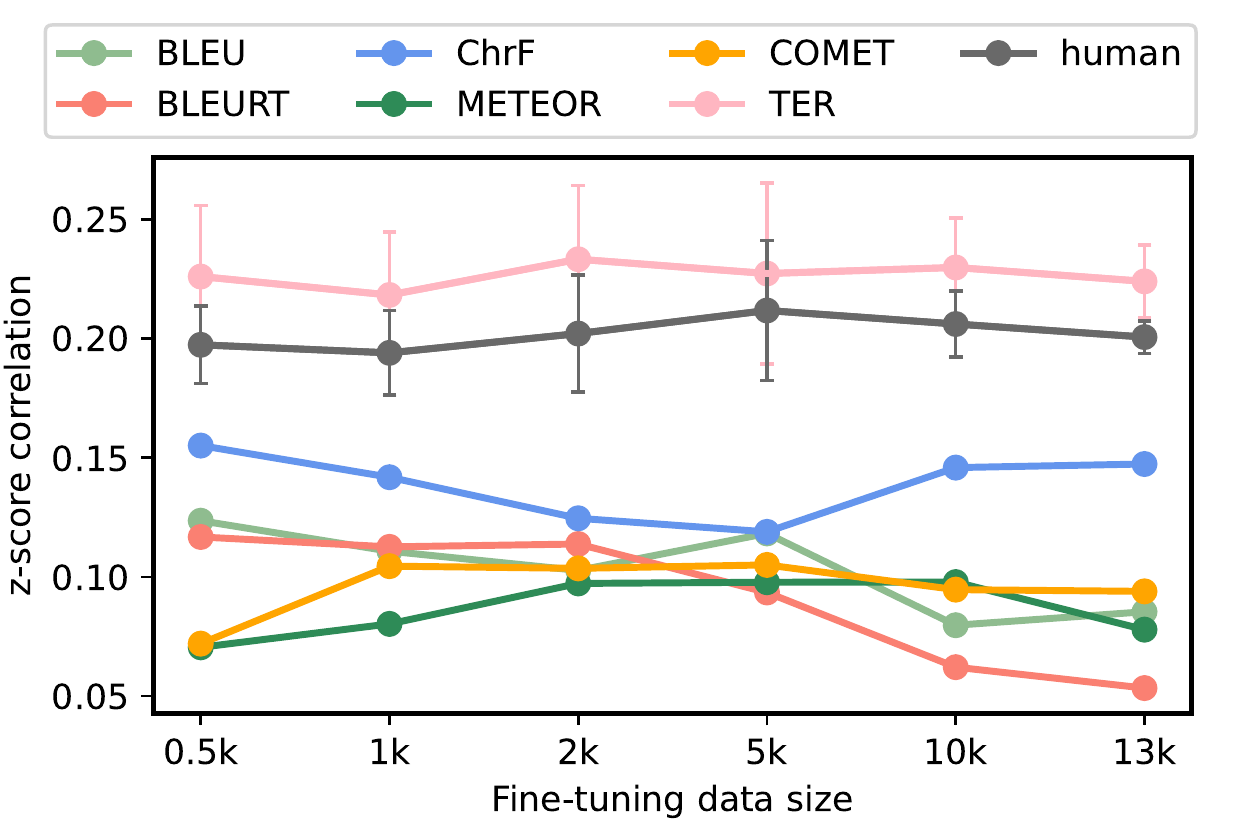}
\caption{Finetuning on limited target-domain QE data. Note the non-linear x-axis. Each point is an average of 5 runs. Error bars show a 95\% confidence interval of the mean. Error bars shown only for {BLEU} and {human} for clarity.}
\label{fig:limited_finetuning}
\end{figure}

\section{Reproducibility}
\label{sec:reproducibility}

We compute metric scores using SacreBLEU \citep{post2018call} with the following signatures:
\begin{itemize}
\item All: \texttt{nrefs:1 | version:2.2.0}
\item sentBLEU: \texttt{case:mixed | eff:yes | tok:13a | smooth:exp}
\item ChrF: \texttt{case:mixed | eff:yes | nc:6 | nw:0 | space:no}
\item TER: \texttt{case:lc | tok:tercom | norm:no | punct:yes | asian:no}
\end{itemize}

For baseline experiments, multilingual version of BERT was used: \texttt{\ttjustify bert-base-multilingual-cased}.
Translations using T5-small are done with the prefix ``\emph{translate English to German:}''.
For other translation models, we used the following models available on torch.hub under the \texttt{pytorch/fairseq} namespace \citep{ott2019fairseq}:
\begin{itemize}[noitemsep]
\item \texttt{dynamicconv.glu.wmt16.en-de}
\item \texttt{conv.wmt17.en-de}
\item \texttt{transformer.wmt16.en-de}
\item \texttt{transformer.wmt18.en-de}
\end{itemize}

\section{Model details}
\label{sec:model_details}

The metric/quality estimation model specifics are shown in \Cref{tab:model_details}.
Additionally, we concatenate all forward and backward hidden states from both LSTM layers.
A non-standard choice was to use sigmoid as the final activation function which worked better than just the linear output.\footnote{This is dissimilar to the baseline model where linear regression worked better than logistic regression even in the case of metrics with bounded output range.}
However, we also rescaled and centered it in case of BLEU, ChrF and METEOR so that scores of 0 and 100 are attainable.
We eventually did not use this in the main experiments so that a single model setup could be used for all metrics.
All models are trained with early stopping of 10 epochs.
The optimization loss is mean squared error.
Our used model is fairly small in comparison to other models, such as those utilizing the Transformer architecture.
This was an intentional choice with respect to the small amount of data used.
The hyperparameters were chosen manually by best practices with respect to final metric correlation (5 trials).

\begin{table}[htbp]
\centering
\resizebox{\linewidth}{!}{
\begin{tabular}{ll}
\toprule
Optimizer & Adam {\small \citep{Kingma2015AdamAM}} \\
Learning rate & $10^{-6}$ \\
Batch size & 10 (0-padded to longest) \\
\midrule
Vocab size (BPE) & 8192 \\
Vocab embedding & 512 \\
LSTM & Hidden state 128 \\
& 2 bidirectional layers \\
LSTM dropout & 20\% inter-layer \\
& 75\% final hidden state \\
Fusion & Concatenate (512 + 6) \\
Linear & 518 $\rightarrow$ 100 \\
Activation & ReLU \\
Linear & 100 $\rightarrow$ 1 (6/7 for multi)\\
\bottomrule
\end{tabular}
}
\caption{Metric/quality estimation model details.}
\label{tab:model_details}
\end{table}

The TF-IDF featurizer in the linear regression TF-IDF baseline uses variable maximum number of features and the best-performing one is chosen.
The search is logarithmical from $2^4$ to $2^{14}$.

\section{Results for other language pairs}
\label{sec:other_langs}

The paper used figures and examples from the English$\rightarrow$German language direction.
To replicate the findings, we translate 500k sentences for the following language directions, models and datasets:\footnote{WMT14 \citep{wmt14}, Opus Paracrawl \citep{opus_paracrawl,paracrawl}, CCMatrix \citep{schwenk2021ccmatrix,fan2021beyond}.}
\begin{itemize}[left=1mm]
\item
    German $\leftrightarrow$ English \texttt{(WMT14)} \\
    \texttt{\small transformer.wmt19.\{de-en,en-de\}}
\item
    German $\leftrightarrow$ Polish \texttt{(opus\_paracrawl)} \\
    \texttt{\small Helsinki-NLP/opus-mt-\{de-pl,pl-de\}}
\item
    Chinese $\leftrightarrow$ English \texttt{(CCMatrix)} \\
    \texttt{\small Helsinki-NLP/opus-mt-\{zh-en,en-zh\}}
\item
    Czech $\leftrightarrow$ English \texttt{(WMT14)} \\
    \texttt{\small Helsinki-NLP/opus-mt-\{cs-en,en-cs\}}
\item
    Russian $\leftrightarrow$ English \texttt{(WMT14)}\\
    \texttt{\small transformer.wmt19.\{en-ru,ru-en\}}
\item
    French $\leftrightarrow$ English \texttt{(WMT14)} \\
    \texttt{\small Helsinki-NLP/opus-mt-\{fr-en,en-fr\}}
\item
    Hindi $\rightarrow$ English \texttt{(CCMatrix)} \\
    \texttt{\small Helsinki-NLP/opus-mt-\{hi-en,en-hi\}}
\end{itemize}

\begin{figure}[htbp]
\includegraphics[width=\linewidth]{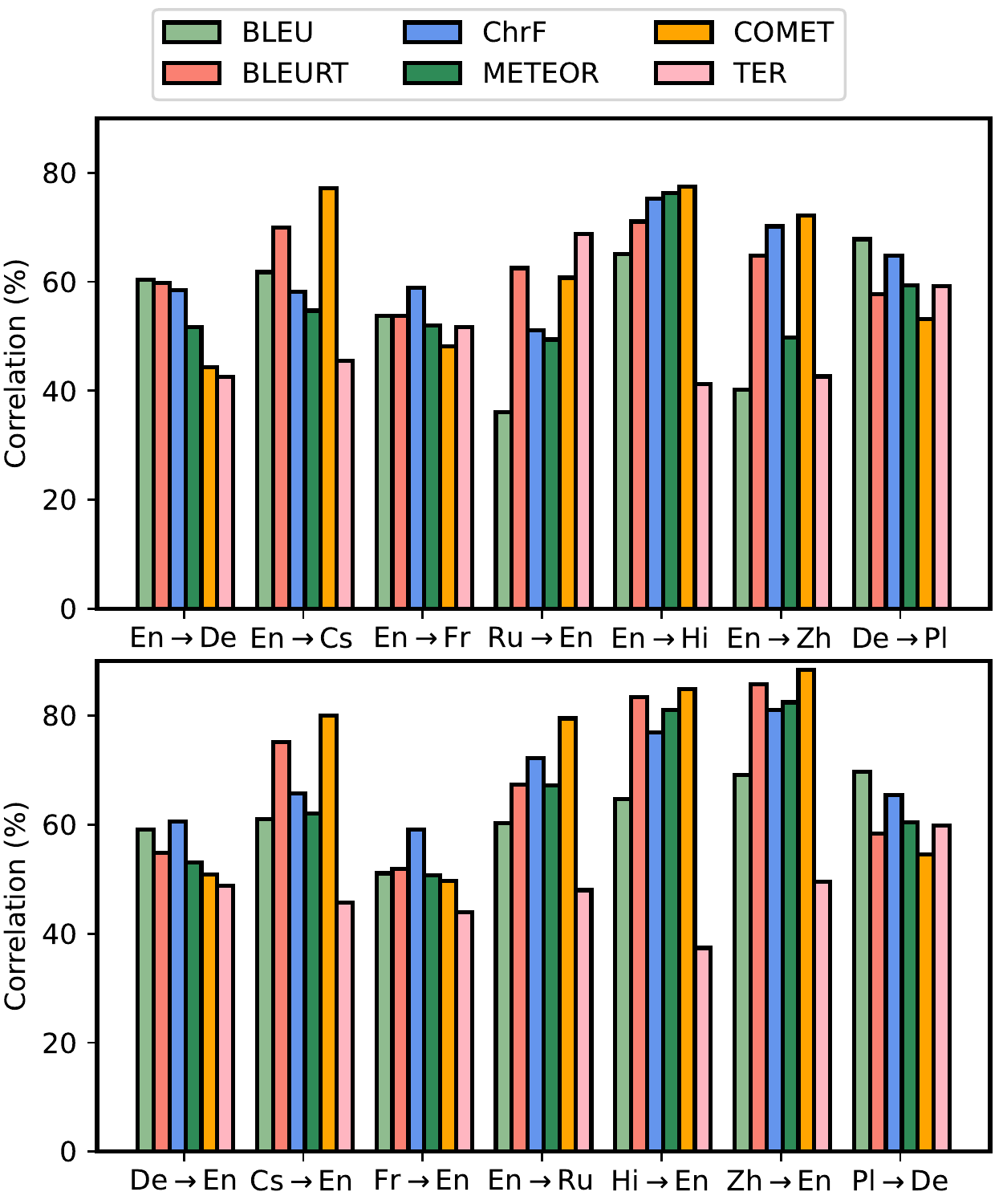}
\caption{Correlations with metrics and human judgement of main metric estimation (ME) models on other languages. Each bar is a separate model trained to predict a particular metric or human judgement.}
\label{fig:baseline_comparison_otherlangs}
\end{figure}

The results for metric estimation for all metrics and language directions are shown in \Cref{fig:baseline_comparison_otherlangs}.
In comparison to the other languages, the chosen language pair in the paper (En $\rightarrow$ De) is more conservative than the other language pairs, which achieve higher correlations across most metrics.
Most of them achieve $>50\%$ correlation, though specifically \emph{COMET} appears to be more predictable for other language pairs.
These results confirm the main results of predictability of the metric without having access to the reference.

\section{Confidence Estimation for Metric Estimation}

Quantifying the confidence of the prediction can be crucial in downstream applications. 
For example, if a quality estimator has low confidence or is predicted to not be accurate, the decision regarding the quality of translation should be delegated to humans.
We attempt to do this by training an auxiliary model to predict the confidence of the output score. 
We do this by training a logistic regression classifier which takes as input the final layer ($\lambda$) of our metric estimator and is trained to predict the binary label: if the predicted metric $f_{\textsc{BLEU}}(s, h, r)$ is close to the true metric $\text{ME}(s, h)$.
\begin{align*}
    \mathbb{P} (|\text{ME} - f_{\textsc{BLEU}} | \in [0, \text{ME}(s, h) \times 10\% ] )  \\
    = \sigma (W^{T} \cdot \lambda + b )
\end{align*}

We, unfortunately, find that the classifier suffers from a very low accuracy 63.4\% against a most common class (negative) baseline of 51.6\%.
It therefore cannot be meaningfully used to ascertain when our regressor is correct and when it is not.
\citet{glushkova2021uncertainty} propose a more complex solution and exploit uncertainty methods for MT metric and quality estimation systems.

\end{document}